\documentclass[sigplan,screen]{acmart}

\usepackage{multirow}
\usepackage{array} 
\usepackage{tcolorbox}

\setcopyright{none}
\renewcommand\footnotetextcopyrightpermission[1]{}
\AtBeginDocument{%
  }

\begin{document}

\title{U-NIAH: Unified RAG and LLM Evaluation for Long Context Needle-In-A-Haystack}

\author{Yunfan Gao}

\affiliation{%
  \institution{Shanghai Research Institute for Intelligent Autonomous Systems, Tongji University}
  \country{China}
}
\email{gaoyunfan1602@gmail.com}

\author{Yun Xiong}
\affiliation{%
  \institution{Shanghai Key Laboratory of Data Science, School of Computer Science, Fudan University}
    \country{China}
    % \country{China}
  % \city{Hekla}
  % \country{Iceland}}
  }
\email{yunx@fudan.edu.cn}

\author{Wenlong Wu}
\affiliation{%
  \institution{College of Artificial Intelligence, Nanjing University of Aeronautics and Astronautics}
    \country{China}
    % \country{China}
  % \city{Rocquencourt}
  % \country{France}
}
\email{wuwenlong@nuaa.edu.cn}

\author{Zijing Huang}
\affiliation{%
 \institution{Shanghai Key Laboratory of Data Science, School of Computer Science, Fudan University}
   \country{China}
   % \country{China}
}
\email{21262010019@m.fudan.edu.cn}

\author{Bohan Li}
\affiliation{%
  \institution{College of Artificial Intelligence, Nanjing University of Aeronautics and Astronautics}
    \country{China}
    % \country{China}
}
\email{bhli@nuaa.edu.cn}

\author{Haofen Wang}
\authornote{Corresponding Author}
\affiliation{%
  \institution{College of Design and Innovation, Tongji University}
    \country{China}
    % \country{China}
    }
\email{carter.whfcarter@gmail.com}

% \author{}
% \affiliation{%

% \author{}
% \affiliation{%

\renewcommand{\shortauthors}{Gao et al.}

\begin{abstract}
Recent advancements in Large Language Models (LLMs) have expanded their context windows to unprecedented lengths, sparking debates about the necessity of Retrieval-Augmented Generation (RAG). To address the fragmented evaluation paradigms and limited cases in existing Needle-in-a-Haystack (NIAH), this paper introduces U-NIAH, a unified framework that systematically compares LLMs and RAG methods in controlled long context settings. Our framework extends beyond traditional NIAH by incorporating multi-needle, long-needle, and needle-in-needle configurations, along with different retrieval settings, while leveraging the synthetic Starlight Academy dataset—a fictional magical universe—to eliminate biases from pre-trained knowledge. Through extensive experiments, we investigate three research questions: (1) performance trade-offs between LLMs and RAG, (2) error patterns in RAG, and (3) RAG’s limitations in complex settings. Our findings show that RAG significantly enhances smaller LLMs by mitigating the ``lost-in-the-middle" effect and improving robustness, achieving an 82.58\% win-rate over LLMs. However, we observe that retrieval noise and reverse chunk ordering degrade performance, while surprisingly, advanced reasoning LLMs exhibit reduced RAG compatibility due to sensitivity to semantic distractors. We identify typical error patterns including omission due to noise, hallucination under high noise critical condition, and self-doubt behaviors. Our work not only highlights the complementary roles of RAG and LLMs, but also provides actionable insights for optimizing  deployments. Code:\url{https://github.com/Tongji-KGLLM/U-NIAH}.
\end{abstract}

\maketitle

\section{Introduction}
% \label{s:introduction}
% \noindent
With the continuous scaling of model parameters and the expansion of context windows to more than 1 million tokens~\cite{long-context_survey}, modern LLMs exhibit remarkable proficiency in processing extended text. This progression has catalyzed a paradigm shift towards long-context (LC) applications, where models are expected to reason over document-level inputs spanning hundreds of pages~\cite{qwen2}. However, the pursuit of extended context windows faces critical challenges in computational efficiency and information retrieval accuracy, especially when handling complex queries in lengthy documents, and research shows that model performance drops rapidly in ultra-long contexts, with the``Lost in the middle" phenomenon occurring~\cite{lost_in_the_mddile}.

In parallel to LC advancements, Retrieval-Augmented Generation (RAG)~\cite{retrieval_survey,zhang2024bench} has already emerged as a prominent alternative architecture that synergizes LLM with external knowledge retrieval. The core debate centers on whether RAG remains essential when LLMs achieve very extended context windows even unlimited context~\cite{li2024long}. The fundamental tension between these approaches lies in their divergent strategies for context processing: while LC models attempt to read all contextual information within their context window, RAG uses an additional retriever to select the most relevant passages from a large document indexing database~\cite{hsieh2024ruler}. Recent empirical studies have demonstrated RAG's superior efficiency in scenarios requiring precise information localization~\cite{oprag}, sparking an ongoing debate about whether LC expansion could potentially obviate the need for retrieval-based method.

Existing research, particularly the Needle-in-a-Haystack (NIAH) evaluation framework\footnote{\url{https://github.com/gkamradt/LLMTest_NeedleInAHaystack}}, has developed standardized methodologies for assessing the performance of LLMs in long-context scenarios. However, current evaluation paradigms face two critical limitations: (1) the lack of unified metrics for cross-paradigm comparison between RAG  and standalone LLM approaches, and (2) the absence of more diverse test cases to further analyze the performance of LLMs in long-context scenarios from multiple dimensions. This methodological fragmentation hinders rigorous comparisons and obscures fundamental insights into the complementary strengths of these architectures. Therefore, establishing a unified evaluation framework is imperative. 

To address the aforementioned limitations, we introduce the Unified Needle-in-a-Haystack (U-NIAH) framework, which unifies the scenarios of LLMs and RAG. The framework extends the capabilities in multiple dimensions, including more complex ``Needle" types, more flexible Needle insertion method , backgrounds with more distractors, and more RAG setting, like retrieval scope and chunk rank. Supporting this framework, we curate Starlight Academy-a meticulously constructed benchmark dataset set in a fictional magical universe, explicitly designed to eliminate knowledge contamination risks. Drawing inspiration from GPQA's ``Google-proof" philosophy~\cite{gpqa}, Starlight Academy ensures that all factual elements are synthetically generated, enabling precise measurement of contextual reasoning capabilities independent of LLMs' prior knowledge~\cite{xiang2024certifiably}.

In constructing the proposed framework, the study lays the groundwork for investigating three interrelated Research Questions (RQ) that illuminate the complementary, substitutive, and developmental relationships between RAG and LLMs within  long-context  tasks. Building on the comparative analysis of both approaches in a needle-in-a-haystack scenario, the first research question (RQ1) examines the distinct contexts in which LLMs and RAG systems demonstrate superior performance under a unified LC setting. The second query (RQ2) seeks to identify and characterize the typical error patterns exhibited by RAG systems when deployed in LC tasks, while the third query (RQ3) explores how RAG systems fare under more challenging and demanding scenarios.

By addressing the three research questions, this study provides a comprehensive overview of the overall performance of RAG and LLM methods in long-context settings, as well as clarifying the specific contexts in which each technique is most effective. The analysis examines the influence of various factors on model performance in LC tasks, systematically identifying recurrent error patterns that offer valuable insights for practical RAG applications. Furthermore, through the design of progressively more challenging scenarios, the study emphasizes the limitations and deficiencies encountered by RAG in LC contexts while also exploring how reasoning models like Deepseek-R1~\cite{deepseek_r1} affect its performance in extended text retrieval tasks. In summary, the main contributions of this paper are as follows: 

\begin{itemize}
    \item We propose a unified evaluation framework U-NIAH, which establishes a mapping between RAG and the original ``needle in a haystack" experiments, thereby providing a unified evaluation benchmark. This framework effectively addresses the fragmentation inherent in traditional evaluations and offers a standardized experimental platform for cross-paradigm comparisons.
    
    \item The study transcends the rudimentary original configurations of NIAH by constructing a dataset based on a completely fictitious background, thereby mitigating interference from the LLM's pre-existing knowledge. It systematically broadens the scope of NIAH's parameters by incorporating multi-needle, long-needle, and nested-needle configurations, in addition to multiple RAG construction settings. The complete code and analysis script are open source, which supports dynamic extension of evaluation scenarios.
    
    \item We investigate the developmental and derivative relationships between LLMs and RAG within LC by formulating corresponding three research questions, analyzing diverse applicable scenarios, and summarizing error patterns. Through more complex experimental settings, we uncover the challenges that current RAG approaches face within LC, thereby guiding future directions.
\end{itemize}

\begin{figure*}[htbp]
    \centering
    \includegraphics[width= \linewidth]{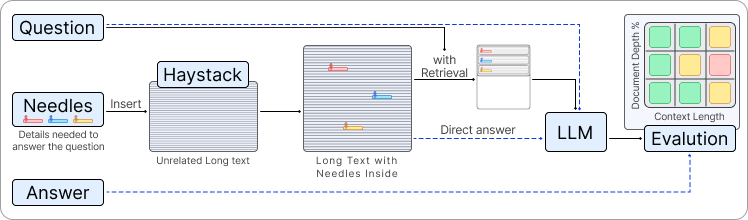}
    \caption{The framework of U-NIAH}
    \label{fig:framework}
\end{figure*}

\section{Related Work}
\noindent
The rapid advancement of LLMs with million-token-scale in-context windows \cite{bulatov2023scaling, ijcai2024p917, basmov2024llms} has fundamentally transformed the paradigm of RAG systems. RAG enhance LC processing capabilities by integrating external knowledge sources, enabling models to manage extensive contexts more effectively~\cite{laban2024summary,gao2024modular}. However, empirical studies \cite{xu2023retrieval, qiu2025eliciting} reveal paradoxical divergences in performance trajectories among different types of models. Commercial models such as GPT-4 demonstrate sustained performance improvements when increasing the number of retrieved passages,(e.g.,64 passages) effectively leveraging the advantages of RAG. In contrast, open-source models exhibit performance degradation beyond critical thresholds, forming significant inverted U-shaped curves. This divergence highlights the limitations of relying solely on in-context window expansion and emphasizes the necessity for systematic investigations into the following key challenges \cite{li2024long, chen2023extending}:optimizing the balance between retrieval granularity and computational overhead, understanding the architectural determinants of LC failure modes, and achieving synergistic integration between RAG and advanced in-context extension strategies.

Recent benchmarking efforts in structured knowledge domains have revealed context-dependent performance patterns~\cite{wang-etal-2024-leave, bai-etal-2024-longbench, bai2024longbenchv2}. In document-intensive scenarios such as multi-document question answering, order-preserving mechanisms that maintain the original document sequences significantly improve F1 scores~\cite{oprag}. Additionally, in processing lengthy Wikipedia articles (e.g., exceeding 64k tokens), global semantic modeling reduces reasoning errors by 15\% \cite{fei-etal-2024-extending}, and LongBench-v2~\cite{bai2024longbenchv2} further validates these findings across various LC tasks. Conversely, emerging paradigms of dynamic knowledge integration, such as multi-stage retrieval pipelines, significantly enhance performance through intermediate reasoning layers~\cite{qiu2025eliciting}. Furthermore, fragmented knowledge environments benefit from LongRAG's inherent optimized chunking strategies, which reduce computational load by 30\% through dual-perspective retrieval~\cite{zhao-etal-2024-longrag}.

Different model families exhibit varying architectural vulnerabilities. Comprehensive evaluations on LongBench-v2 reveal that while commercial LLMs outperform open-source alternatives, they still face challenges with contexts exceeding 100k tokens. Open-source models, on the other hand, experience notable performance declines beyond 32k tokens due to decreased ability in following instructions~\cite{leng2024long}. These challenges have spurred innovative mitigation strategies, including hybrid retrieval reranking~\cite{li-etal-2024-retrieval} and LC-aware fine-tuning approaches~\cite{longlora}. Notably, expanding retrieval units to over 4K tokens offers dual benefits—reducing interference and providing performance compensation for less capable models \cite{jin2024long}.

% 扩展上下文处理与RAG架构之间的协同作用开启了有前景的优化前沿。在多语言基准测试中，联合策略通过多模态知识融合实现了显著的增益\cite{oprag}。此外，检索单位数量与上下文窗口大小之间的最优比例\cite{zhao-etal-2024-longrag}的识别为系统设计原则提供了数学上的严谨性。当结合工程指标时，这一发现具有实际意义——优化配置在保持高检索召回率的同时，减少了资源消耗\cite{xu2024chatqa}。

% 评估方法学也相应地发展以应对这些复杂性。LongBench框架\cite{bai-etal-2024-longbench}通过21个双语数据集引入了多维评估，开创了长文本理解的综合指标。专门的模型展示了基于检索压缩的有效性\cite{xu2023retrieval}，而混合方法通过智能查询路由实现了成本效益高的性能\cite{li-etal-2024-retrieval}。这一向系统化评估的范式转变不仅实现了与人工判断的一致性，还支持了实时性能监控——这是生产系统的关键进展\cite{wang-etal-2024-leave}。

The emergent synergy between extended context processing and RAG architectures opens promising optimization frontiers. In multilingual benchmark tests, joint strategies achieve significant gains through multimodal knowledge fusion~\cite{oprag}.  Moreover, specialized models demonstrate the efficacy of retrieval-based compression \cite{xu2023retrieval}, while hybrid approaches achieve cost-effective performance through intelligent query routing~\cite{li-etal-2024-retrieval}. Additionally, identifying the optimal proportion between the number of retrieval units and LC window size~\cite{zhao-etal-2024-longrag} provides mathematical rigor to system design principles. This finding gains practical significance when contextualized with engineering metrics—optimized configurations reduce resource consumption while maintaining high retrieval recall~\cite{xu2024chatqa}.

Evaluation methodologies have correspondingly evolved to address these complexities. LongBench\cite{bai-etal-2024-longbench} introduces multidimensional assessment through 21 bilingual datasets, pioneering comprehensive metrics for LC comprehension, while $\infty$Bench\cite{zhang-etal-2024-bench} further proposes the first standardized benchmark specifically designed for evaluating 100K+ token contexts through synthetic and realistic multilingual tasks, revealing critical performance limitations in existing LC LLMs. This paradigm shift towards systematic evaluation not only achieves consistency with human judgment but also supports real-time performance monitoring—a crucial advancement for production systems~\cite{saad2024benchmarking}.

\section{Framework}
\noindent
The persistent debate on long-context processing between LLMs and RAG motivates our development of the U-NIAH framework. This chapter will first introduce the core content of the framework. Subsequently, based on this framework, we propose the three research questions that this paper attempts to address. Thereafter, in response to the aforementioned research questions, we will provide a detailed introduction to the core dimensions and settings within the framework.

\subsection{A unified NIAH framework}
\noindent
We propose a unified framework that integrates both RAG and LLM for the Needle-in-a-Haystack (NIAH) task. As illustrated in Figure~\ref{fig:framework}, the framework processes an input quadruple $(Q, A, \mathcal{N}, \mathcal{H})$ where $Q$ denotes the query, $A$ the ground-truth answer, $\mathcal{N}=\{n_1,...,n_k\}$ the set of essential information needles, and $\mathcal{H}$ the haystack corpus containing ( irrelevant)  long text. Let $L$ represent the total context window size, with $P$ tokens reserved for system prompts and $N=\sum_{i=1}^k |n_i|$ tokens allocated for needles. The remaining capacity $H = L - P - N$ is filled with haystack text sampled from $\mathcal{H}$. 

\subsubsection{Needle Insertion}
\noindent
The needle insertion process follows a depth-controlled distribution. Given document depth $L$ (token length), the first needle $n_1$ is inserted at a specified relative position $d_1 = \alpha H$ (e.g., $\alpha=10\%$). Subsequent needles $\{n_2,...,n_k\}$ are uniformly distributed in the remaining document space $[d_1, H]$. Specifically, the remaining $H - d_1$ tokens are divided into $k$ equal segments, and the $i$-th needle ($i \geq 2$) is placed at the boundary of the $(i-1)$-th segment. Formally, the absolute insertion positions are computed as:  
$$
\{d_i\}_{i=2}^k = \left\{ d_1 + \left\lfloor \frac{(i-1)(H - d_1)}{k} \right\rfloor \mid i=2,...,k \right\}.
$$
For example, assume the total remaining context length  $H$ tokens and $k=3$ needles. Let the first needle $n_1$ be inserted at $d_1 = 10\%$. The subsequent needles are inserted at: 40\% of $H$ and 70\% of $H$.

This generates an augmented context $C = \text{Insert}(\mathcal{H}_H, \mathcal{N}, \{d_i\})$ that combines haystack content with strategically placed needles. This distribution ensures the needles are spaced through the middle-to-late sections of the document, avoiding clustering near the head or tail.

\subsubsection{Context Construction}
\noindent
For context construction, we formalize two operational modes. The direct LLM approach (LC-LLM) processes the full context directly:
$$\hat{A}_{\text{LLM}} = f_{\theta}(Q,C)$$
where $f_{\theta}$ represents the LLM generation function. The RAG variant (LC-RAG) implements a baseline retrieval pipeline:
\begin{align*}
\mathcal{C} &= \text{Chunk}(C, \tau) \\
\mathcal{E} &= \{E(c_i) | c_i \in \mathcal{C}\} \\
\mathcal{R} &= \text{TopK}(E(Q), \mathcal{E}, k) \\
\hat{A}_{\text{RAG}} &= f_{\theta}(Q,\oplus(\mathcal{R}))
\end{align*}
where $\tau$ denotes chunk size, $E(\cdot)$ the embedding function, $\oplus$ the context aggregation operator, and $\text{TopK}(\cdot)$ the similarity-based retrieval function.

\subsubsection{Evaluation}
\noindent
The evaluation metric $\mathcal{M}: (A, \hat{A}) \mapsto \{1,3,5,7,10\}$ employs an LLM-as-judge scoring mechanism with discrete ordinal levels corresponding to answer quality, where higher scores require stricter adherence to semantic equivalence rather than lexical similarity.  
\begin{itemize}
    \item Score 1: The answer is completely unrelated to the reference or totally wrong.
    \item Score 3: The answer has minor relevance but does not align with the reference.
    \item Score 5: The answer has moderate relevance but contains inaccuracies.
    \item Score 7: The answer partially aligns with the reference but has minor omissions.
    \item Score 10: The answer is aligned with the reference. No need to consider the information order and no need to be exact the same as the reference.
\end{itemize}

For each context length $L \in \mathcal{L}$ and needle depth configuration $\{d_i\}\in \mathcal{D} $, we compute the performance matrix $M \in \mathbb{R}^{|\mathcal{L}| \times |\mathcal{D}|}$ where $\mathcal{D}$ represents the depth parameter space. The resultant heatmap visualization $H = \phi(M)$ reveals the sensitivity patterns of LLM and RAG architectures to contextual variations, with:
$$
H_{ij} = \mathbb{E}[\mathcal{M}(A,\hat{A}) | L_i, d_j]
$$
capturing the expected score at specific length-depth configurations. This unified framework enables systematic comparison of information retrieval capabilities under controlled noise conditions.

\begin{figure*}[htbp]
    \centering
    \includegraphics[width= \linewidth]{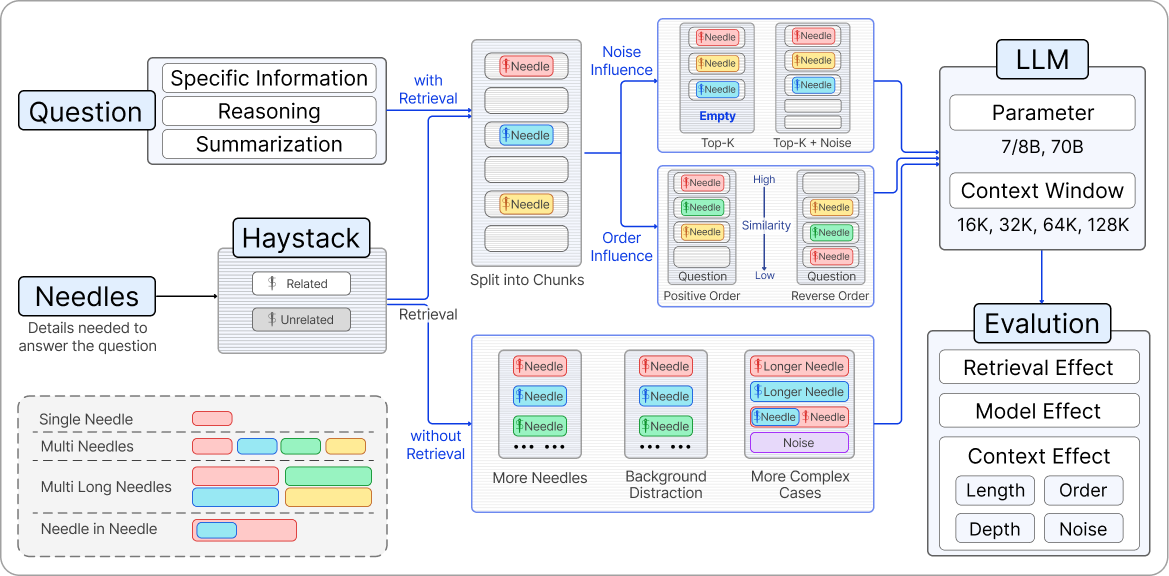}
    \caption{The RAG factors in U-NIAH}
    \label{fig:Rag_factors}
\end{figure*}

\subsection{Research Questions}
\noindent
Based on the U-NIAH, this study focuses on the complementary, substitutable, and developmental relationships between RAG and LLMs. Especially, we propose the following research questions (RQs):

\begin{tcolorbox}[colback=white,colframe=gray!,title={\textbf{RQ1:} How do LLMs and RAG perform, and in which scenarios are they respectively more suitable?}]
This RQ combines the foundational analysis of the current capabilities of LLMs in retrieving and understanding long texts with a comparative analysis of their performance in specific scenarios. Comparing performance of models with different levels of capability and varying parameter sizes to provide a comprehensive overview of their strengths and weaknesses.
\end{tcolorbox}

\begin{tcolorbox}[colback=white,colframe=gray!,title={\textbf{RQ2:} What are the typical error patterns of RAG in the LC scenario?}]
This question focuses on error attribution and summarizing the common error patterns of RAG in different scenarios. By identifying these patterns, we can gain insights into the challenges and limitations of RAG in handling complex long-text tasks, which will inform the development of more robust solutions.
\end{tcolorbox}

\begin{tcolorbox}[colback=white,colframe=gray!,title={\textbf{RQ3:} How does RAG perform when facing more challenging scenarios? In which aspects are its limitations reflected?
}]
This question explores the potential directions for further optimization of RAG. By increasing the difficulty of the test scenarios, we aim to identify the critical factors that need to be addressed to enhance the capabilities of RAG in handling more challenging tasks. This will provide valuable guidance for future research and development efforts in this area.
\end{tcolorbox}

\begin{figure*}[htbp]
    \centering
    \includegraphics[width=1\linewidth]{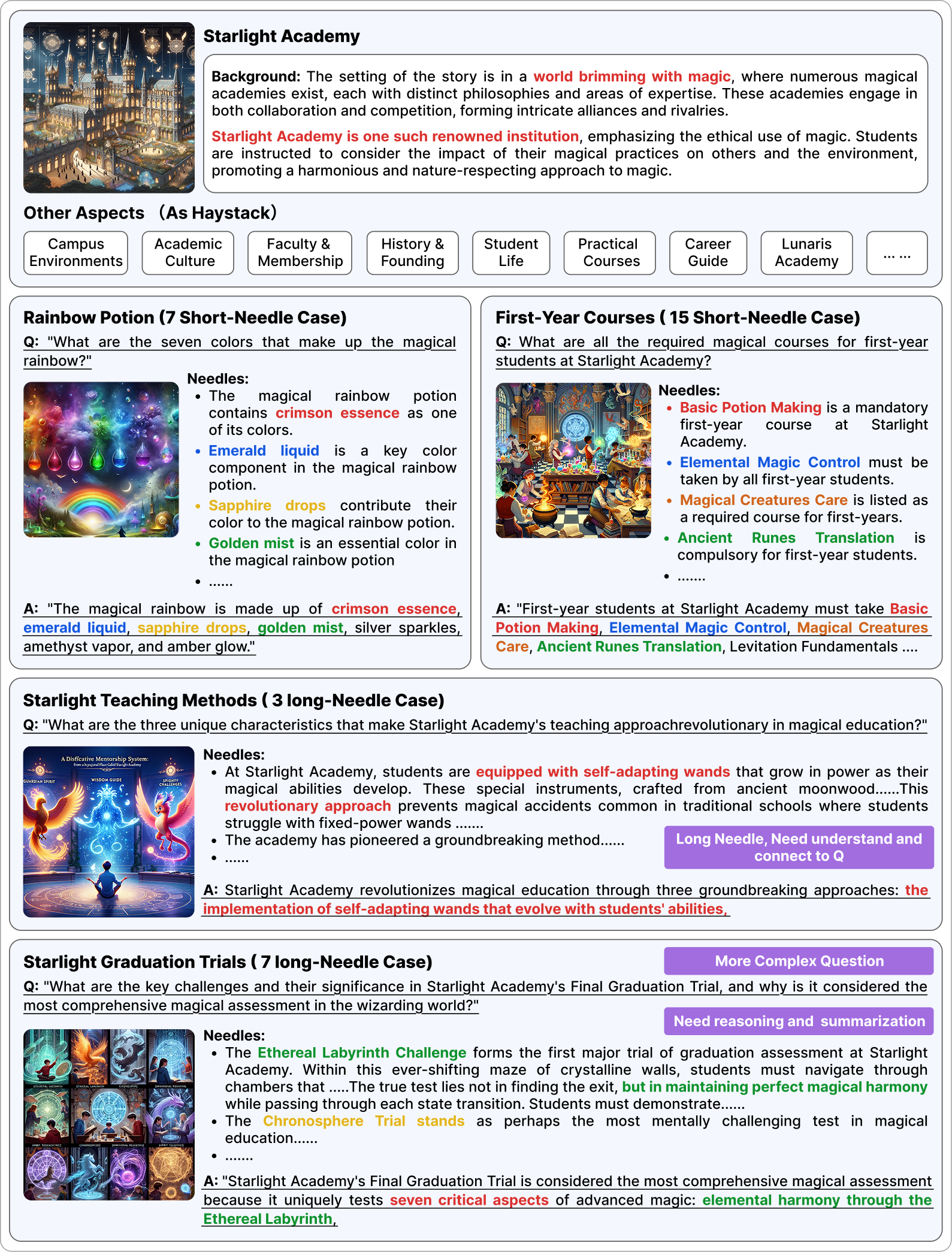}
    \caption{The dataset of Starlight Academy}
    \label{fig:Starlight}
\end{figure*}

\subsection{ Multidimensional Factors}
\noindent
To comprehensively address the research questions and simulate real-world RAG scenarios we extend the original NIAH paradigm and define the following key factors.

\textbf{Needle}. We extend the original NIAH paradigm along three critical axes: 
\begin{enumerate}
    \item Quantity: The framework supports multi-needle scenarios with configurations of 3, 7, and 15 needles, substantially increasing the information retrieval challenge. 
    \item Length: Needle lengths vary from concise single-sentence statements (50-100 tokens) to extended passages-level needls, which are consist of 4-5 sentences (400-500 tokens).
    \item Composition: To further escalate the difficulty, we introduce a hierarchical "Needle-in-Needle" structure where short critical needles are embedded within longer passage-level needles.
\end{enumerate}

\textbf{Query}. Beyond simple and detailed fact retrieval queries, our framework incorporates more complex questions that require: (1) accurate needle identification in extended contexts, (2) cross-needle relationship analysis, and (3) synthetic summarization capabilities. Answer formats correspondingly extend from single-sentence extracts to structured multi-paragraph responses, particularly in scenarios with long-form needles.

\textbf{Haystack}. While maintaining the baseline setting of semantically unrelated background text $\mathcal{H}$ , we introduce adversarial haystack $\mathcal{H}'$ configurations containing distractors – text segments that share semantic similarity with needles but do not contain accurate information . This enhancement increases the framework's realism by simulating real-world scenarios where relevant-looking distractors may mislead retrieval systems.

\textbf{Context Construction.} Our framework introduces two critical augmentation dimensions for context build during  RAG pipeline.

1) Noise ratio $\eta = \frac{|\mathcal{R}_{\text{noise}}|}{|\mathcal{R}|}$ controlled through:  
$$\mathcal{R}' = \mathcal{R}_{\text{topK}} \cup \mathcal{H}_{\text{rand}}^{\lfloor \eta L \rfloor}$$  
First, noise injection is implemented through controlled retrieval expansion: beyond the standard top-k chunks, we fill  more (e.g. ,50\%, 100\%) of the remaining context window capacity with lower-relevance chunks. On the one hand, this allows for a more equitable basis of comparison for LLMs. On the other hand, it also aligns more closely with the current application scenarios of LLM-related applications. For example, using LLMs to process massive amounts of information retrieved from the web.

2) Context ordering. Second, we formalize two chunk ordering schemes – Norm (descending relevance order) and Reverse (ascending relevance order) – to investigate positional effects. Formally, given retrieved chunks $\mathcal{R} = \{r_1,...,r_k\}$ sorted by relevance scores $\{s_1 \geq ... \geq s_k\}$, the context aggregation operator $\oplus$ implements:
$$
\oplus(\mathcal{R}) = \begin{cases}
\langle r_1 \parallel ... \parallel r_k \rangle & \text{(Norm)} \\
\langle r_k \parallel ... \parallel r_1 \rangle & \text{(Reverse)}
\end{cases}
$$
Previous research on prompt engineering has already shown that placing key information closer to the query helps LLMs to better utilize the information~\cite{white2023prompt}. To verify whether this conclusion is also applicable under LC, This creates distinct needle distributions relative to the query position, enabling systematic analysis of proximity effects in long contexts.

\textbf{Model Considerations}. We systematically examine the impact of generation variations through three model dimensions: parameter scale (7B to 70B) and context window size (16k-128k tokens). This multi-axis comparison aims to demonstrate whether the performance of current LLMs in NIAH is affected by model parameters and model tier (e.g., GPT-4o and GPT-4o-mini). It also seeks to identify if there are specific behavioral patterns among different model families. The results reveal potential interactions between model capacity, context length utilization, and inherent model capabilities.

\section{Dataset and Experiment Setting}
\label{s:Experimental}
\noindent
\subsection{Dataset}
\noindent
The Needle-in-a-Haystack (NIAH) experiment is designed to evaluate LLMs' capability to retrieve specific information from extended contexts while minimizing interference from their prior knowledge. To achieve this objective, we constructed an entirely fictional magical world framework centered around Starlight Academy - a premier institution for arcane education. This synthetic universe was collaboratively developed through iterative generation by multiple LLMs  to ensure content diversity and mitigate potential biases from any single model. The comprehensive world-building encompasses detailed descriptions of magical systems, academic curricula, campus life, teaching methodologies, and inter-institutional relationships with its sibling school Lunaris Academy. Within this controlled narrative environment, we designed four distinct NIAH evaluation dataset with varying complexity scenarios, which is illustrated in Figure~\ref{fig:Starlight}.

This section presents concrete case studies demonstrating various evaluation scenarios. A representative example is the Rainbow Potion 7 short-needle case. The test requires identifying seven distinct elements, where each element combines specific chromatic attributes with material components. Successful resolution demands accurate extraction of all color-substance pairs without omission or hallucination.

Another illustrative case involves First-year Course Identification, which is 15 short-needle case. The task presents a comprehensive curriculum listing 15 mandatory courses for Starlight Academy freshmen, alongside distractor items from subsequent academic years. Effective performance necessitates precise recognition of all target courses while maintaining resistance to irrelevant information from upper-year programs.

The Starlight Teaching Method scenario constitutes a 3 long-needle case. Each approximately 500 token needle describes unique pedagogical approaches employed at the academy. Beyond mere information retrieval, this test requires models to establish connections between extended textual descriptions and their corresponding educational characteristics, thereby evaluating both comprehension and relational reasoning abilities.

For complex synthesis challenges, the Starlight Graduation Trials 7 long-needle case presents a multifaceted evaluation. This scenario requires models to not only locate dispersed examination requirements across extended contexts but also synthesize fragmented information to address explanatory queries (typically ``why"-type questions). Successful resolution demonstrates advanced capabilities in information integration, causal reasoning, and coherent summarization of distributed narrative elements. In addition, we still retain the three-needle case of Pizza Ingredients from the original NIAH to facilitate comparison with the original framework.

\begin{figure*}[htbp]
    \centering
    \includegraphics[width=1\linewidth]{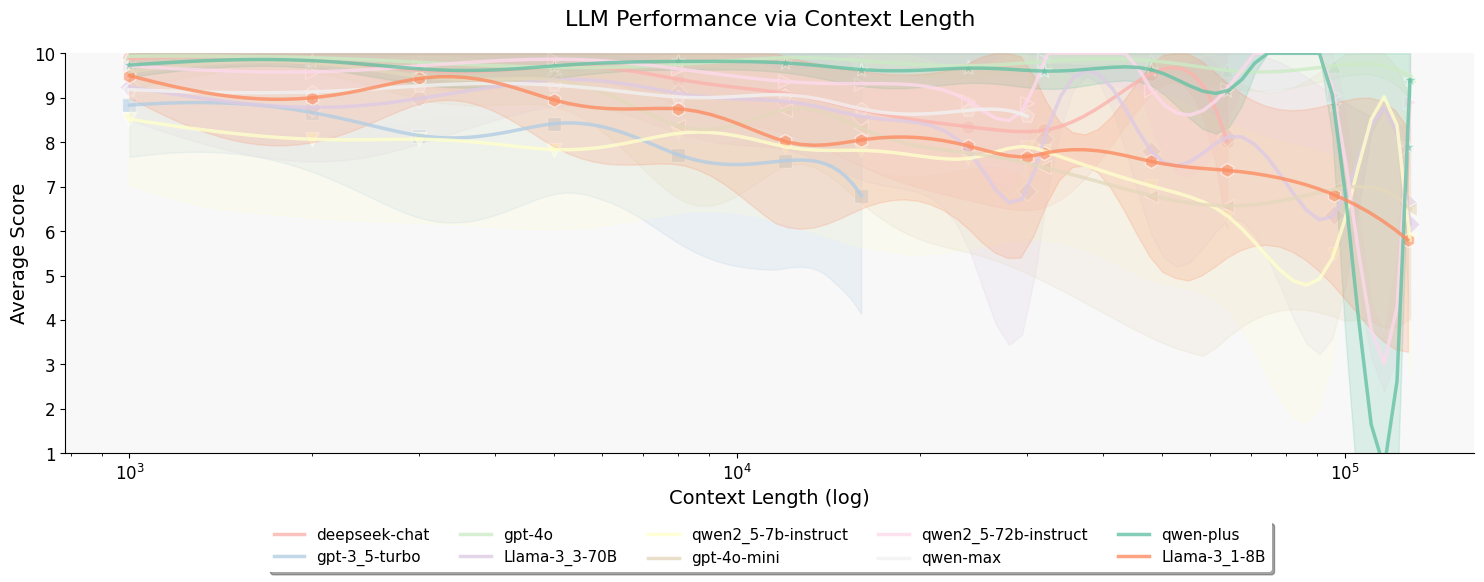}
    \caption{LLM Performance via context length in U-NIAH Framework}
    \label{fig:LLM_LC}
\end{figure*}

\subsection{Experiment Setting}
\noindent
For experimental setting, we established context windows ranging from 1k to 128k tokens with needle insertion positions systematically distributed from 10\% to 100\% depth (10\% increments). When the insertion depth is 100\%, all Needles are inserted directly before the question at the very bottom of the Context. To maintain syntactic coherence, needle insertion respected sentence boundaries, resulting in actual context lengths slightly shorter than nominal values. The evaluation encompassed state-of-the-art LLMs spanning multiple architectural families (OpenAI GPT-series, DeepSeek-v3, Qwen2.5 series , LLaMA3 series) with temperature parameters fixed at 0 (or 0.1 for models requiring minimum non-zero settings). 

The RAG implementation employed OpenAI's text-embedding-3-small model with fixed chunking parameters (600-token windows, 100-token overlap) to isolate retrieval mechanism variations. This configuration focuses specifically on assessing RAG-enhanced LLMs' needle detection capabilities rather than optimizing individual retrieval components. For the case with 3, 7, and 15 needles, under the TopK setting, we respectively retrieve the top 5, 10, and 20 chunks, leaving a certain amount of redundancy.

\section{Results and Discussion}
\subsection{Performance of RAG and LLM in U-NIAH }
\noindent
To address RQ1 regarding the comparative performance of LLMs and RAG systems in needle-in-a-haystack  tasks under unified long-context (LC) settings, we systematically evaluated 10 LLMs across scenarios containing 3, 7, and 15 short/long needles. 

\begin{figure*}[ht]
    \centering
    \includegraphics[width=1\linewidth]{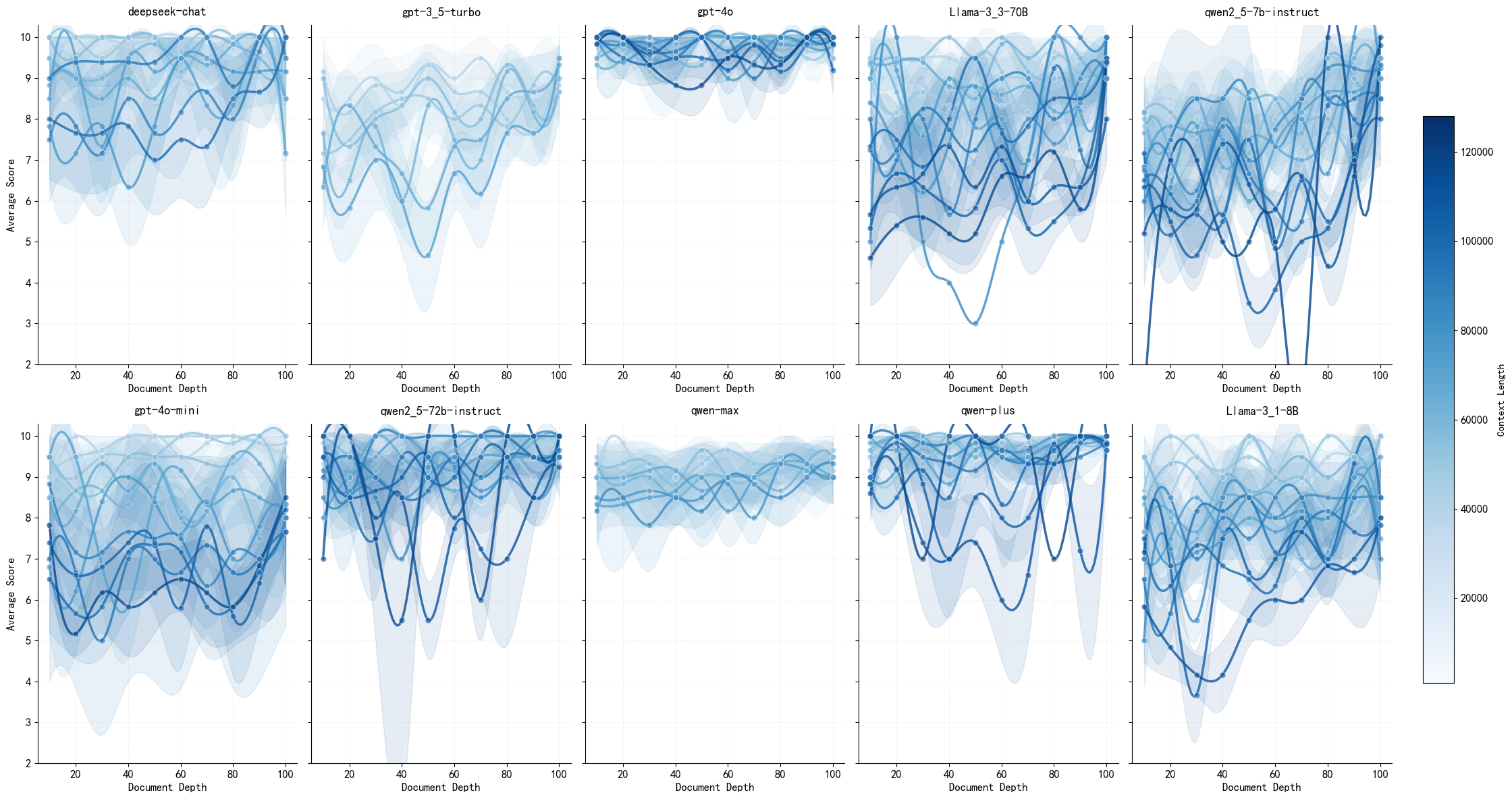}
    \caption{Performance of LLM in U-NIAH framework under different Document Depth}
    \label{fig:LLM_depth}
\end{figure*}

\subsubsection{LLM Performance}
\noindent
The performance of pure LLM is illustrated in Figure~\ref{fig:LLM_LC}, the trends reveal critical insights into model capabilities as context length scales. The horizontal axis represents context length (ranging from 1K to 128K tokens), while the vertical axis denotes average retrieval scores calculated by aggregating results across all doument depth and cases. Shaded error bands indicate ±1 standard deviation from the mean. The varying maximum context lengths of different models result in some curves terminating in the middle of the graph. For instance, the maximum length of GPT-3.5-turbo is 16K, while the maximum context length of DeepSeek-V3 is 64K~\cite{deepseek_v3}.

\begin{figure*}[h]
    \centering
    \begin{minipage}[b]{0.32\textwidth}
        \centering
        \includegraphics[width=\textwidth]{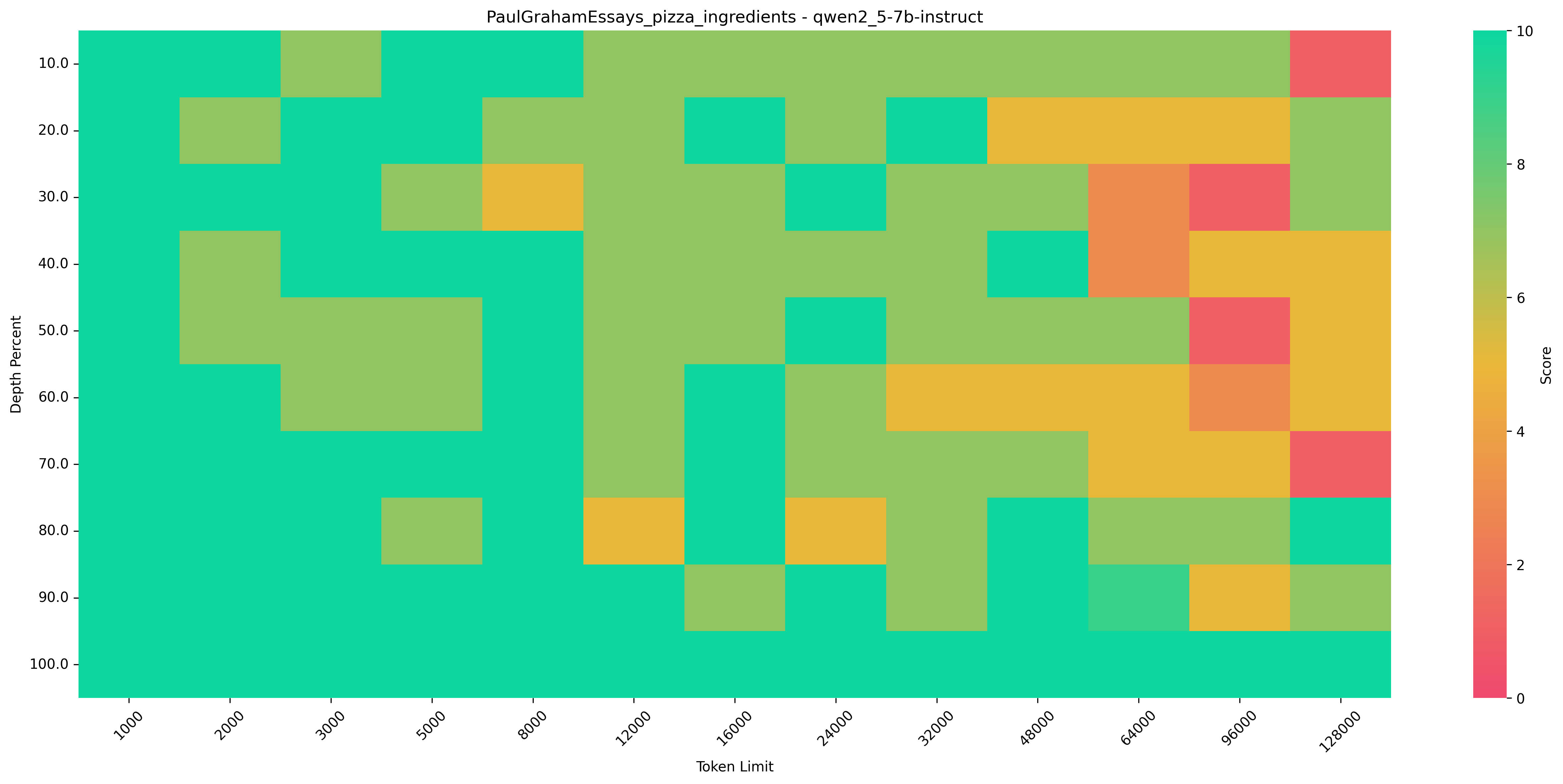}
        \centerline{(a) 3 Needles Qwen2.5-7B}
    \end{minipage}
    \hfill
    \begin{minipage}[b]{0.32\textwidth}
        \centering
        \includegraphics[width=\textwidth]{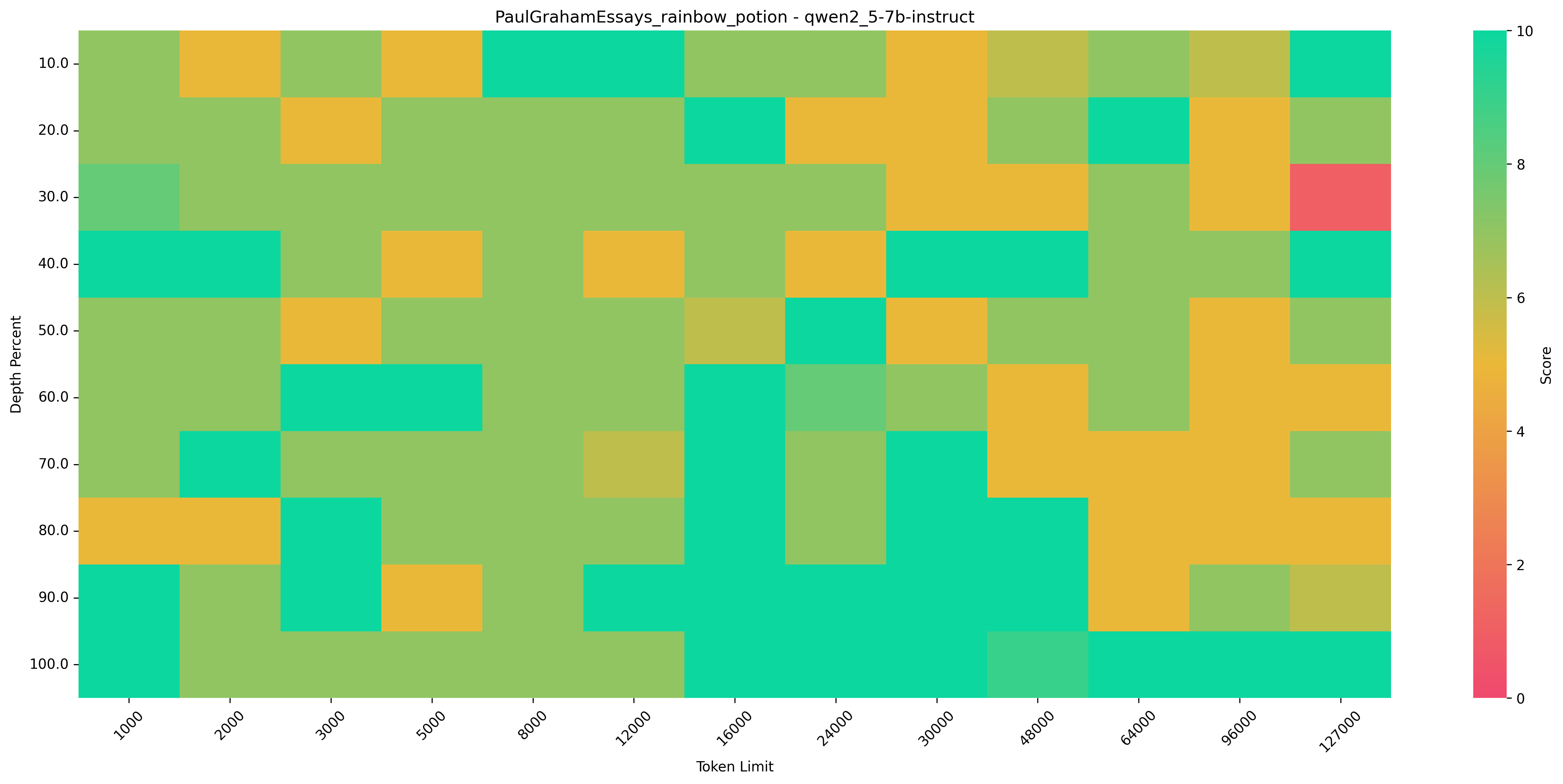}
        \centerline{(b) 7 Needles Qwen2.5-7B}
    \end{minipage}
    \hfill
    \begin{minipage}[b]{0.32\textwidth}
        \centering
        \includegraphics[width=\textwidth]{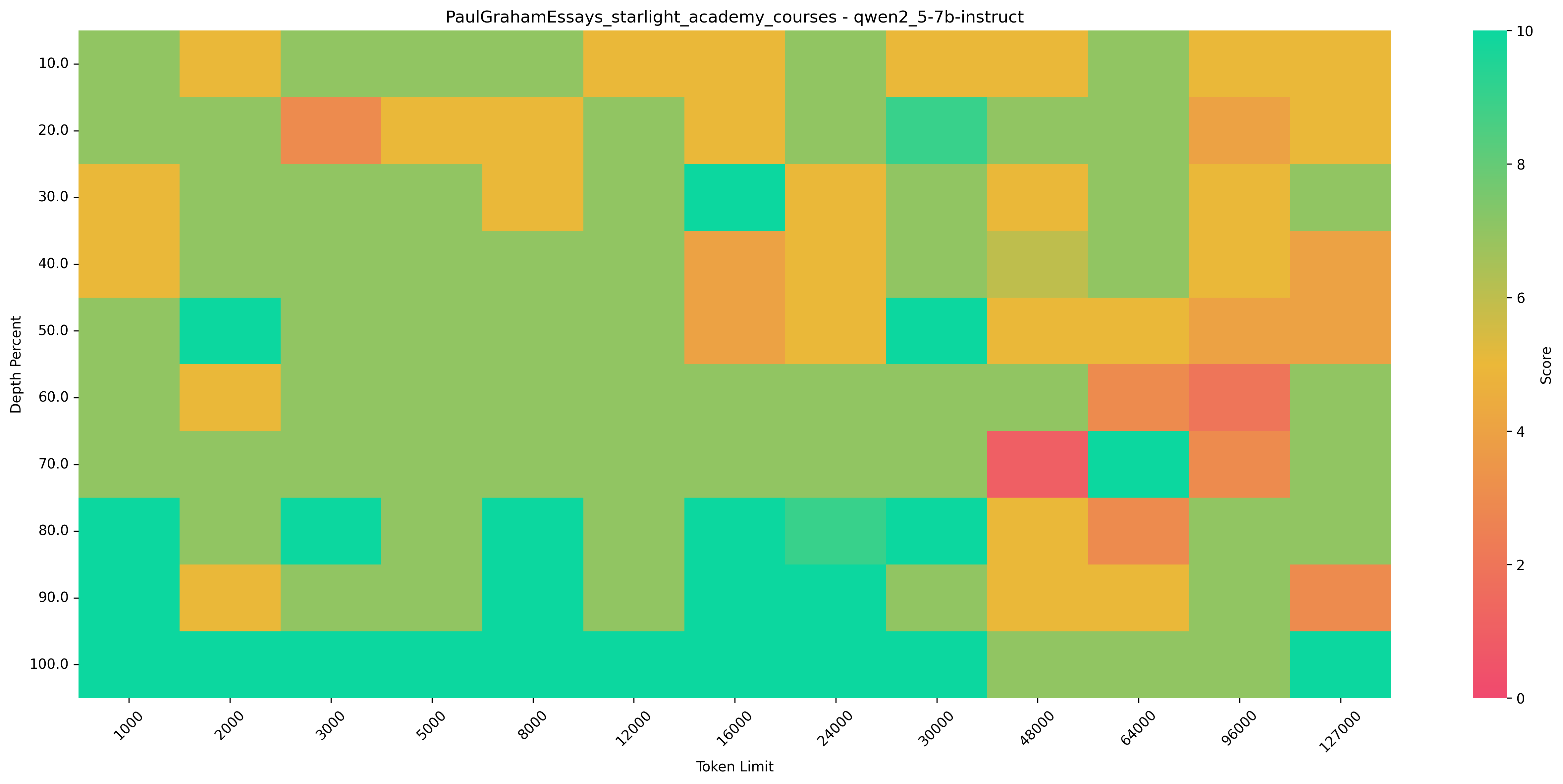}
        \centerline{(c) 15 Needles Qwen2.5-7B}
    \end{minipage}
    
    \begin{minipage}[b]{0.32\textwidth}
        \centering
        \includegraphics[width=\textwidth]{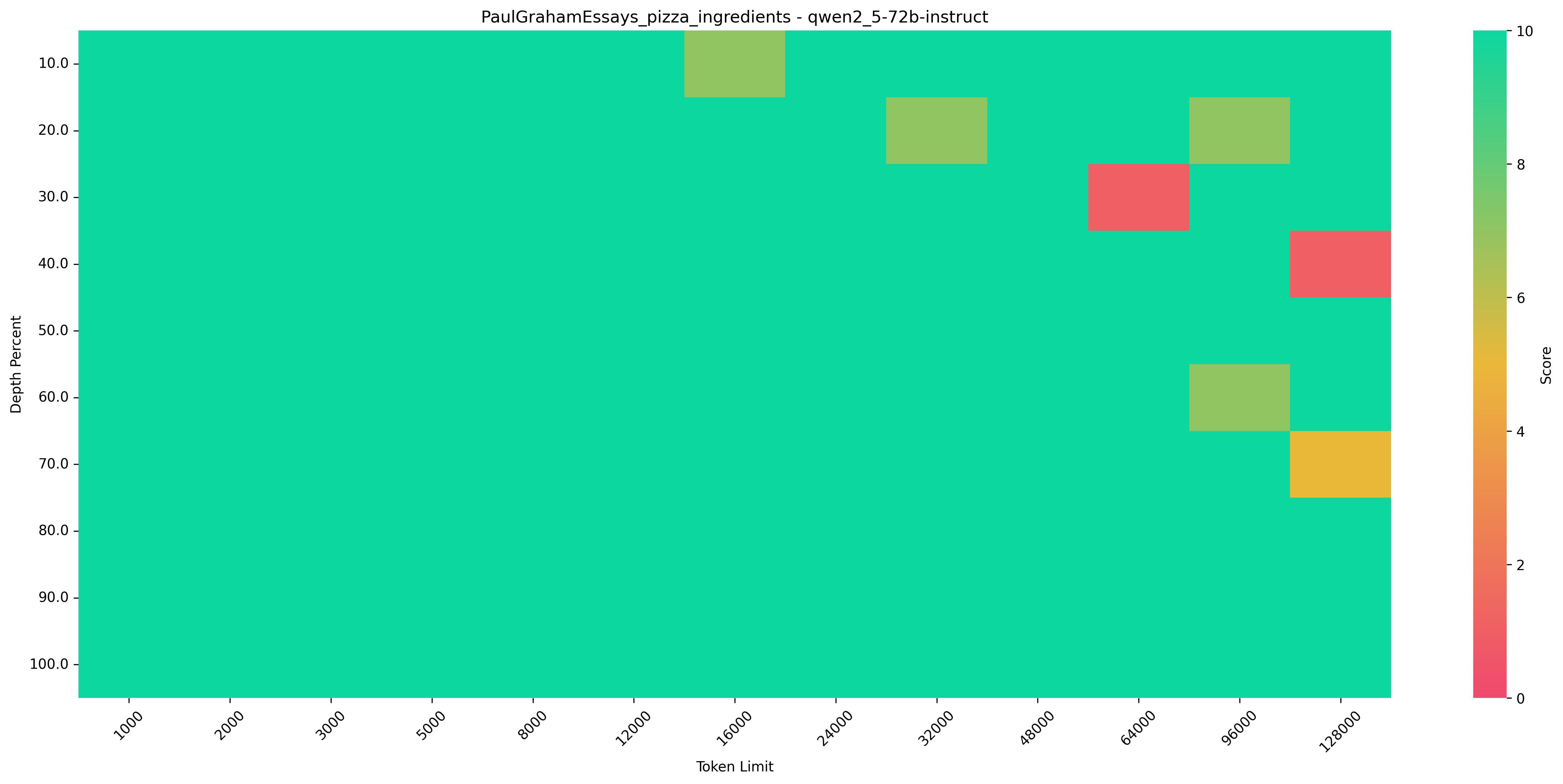}
        \centerline{(d) 3 Needles Qwen2.5-72B}
    \end{minipage}
    \hfill
    \begin{minipage}[b]{0.32\textwidth}
        \centering
        \includegraphics[width=\textwidth]{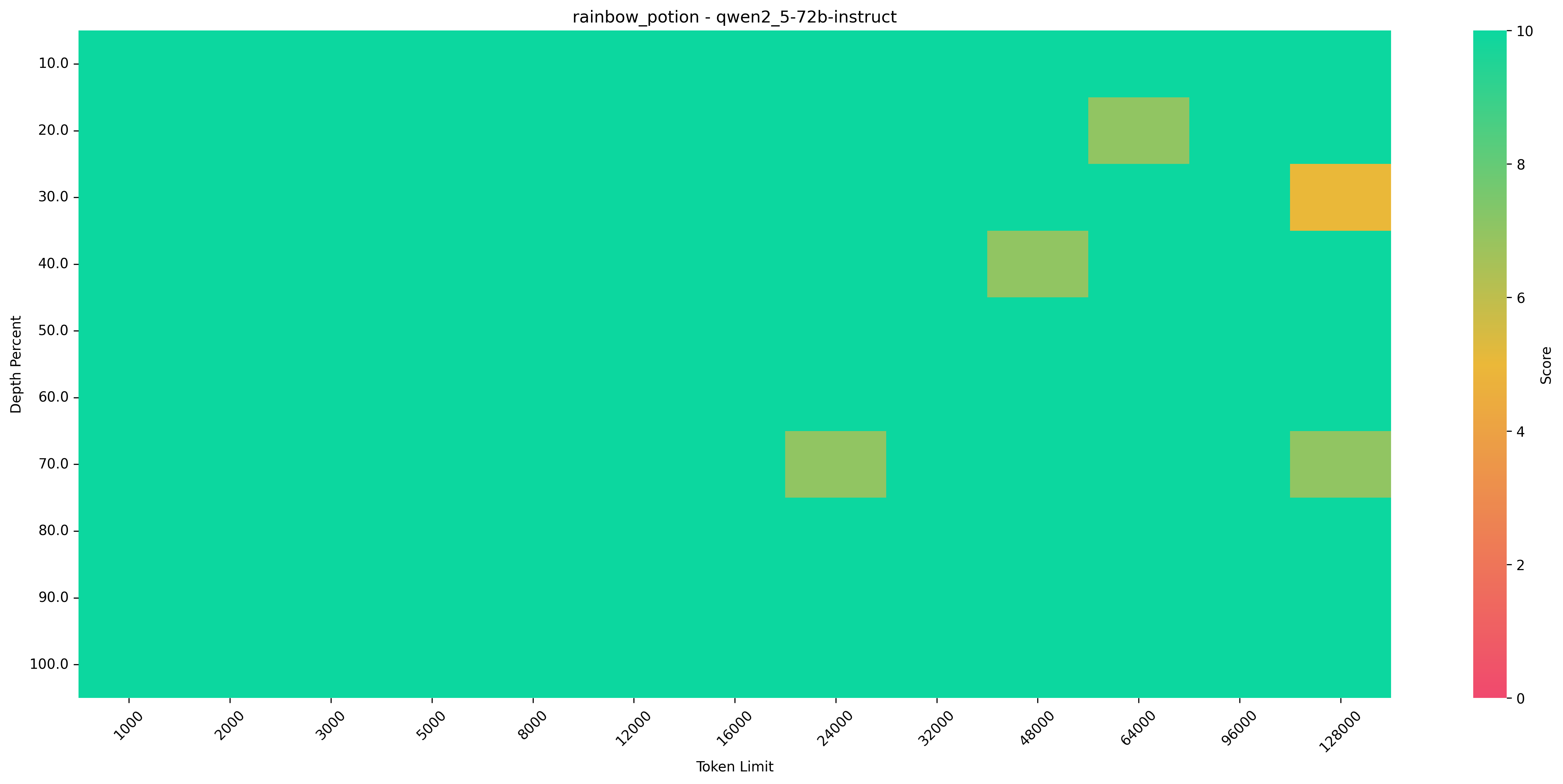}
        \centerline{(e) 7 Needles Qwen2.5-72B}
    \end{minipage}
    \hfill
    \begin{minipage}[b]{0.32\textwidth}
        \centering
        \includegraphics[width=\textwidth]{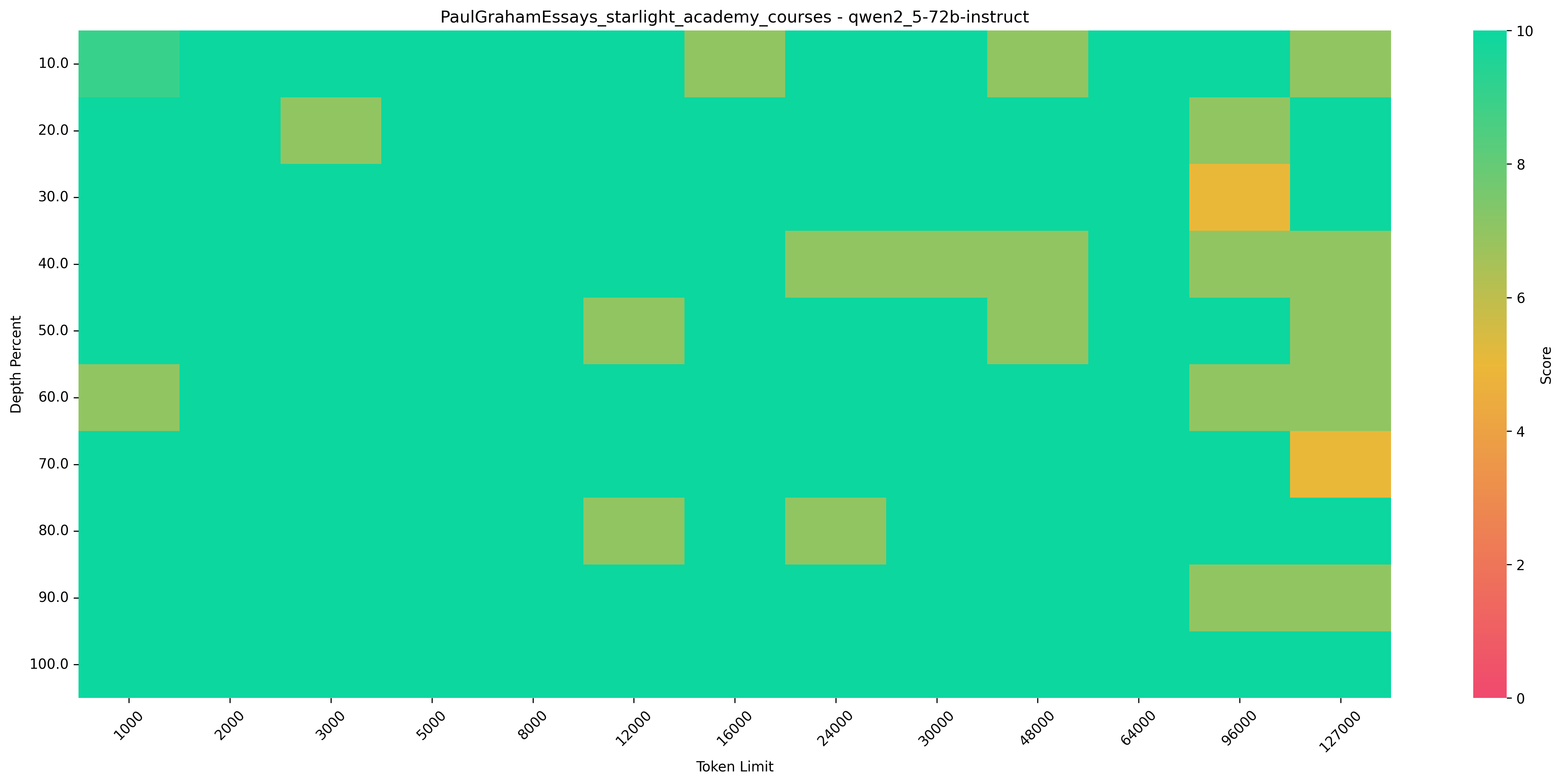}
        \centerline{(f) 15 Needles Qwen2.5-72B}
    \end{minipage}
    \caption{Comparison of heatmap between Qwen2.5-7B and Qwen2.5-72B}
    \label{fig:qwen_compare}
\end{figure*}

In all the experiment settings, the LLM achieved an overall score of 8.67 with a standard deviation of 2.01. It can be observed that the performance of the model fluctuated with the increase of the context length. In shorter context windows, the model demonstrated relatively stable performance. However, as the context window became larger, the model's performance began to decline and exhibited greater fluctuations. The widening of the error band indicated that the LLM still had inherent instability in the NIAH task with long contexts. In terms of model dimensions, models with larger parameter scales and higher levels (e.g. GPT-4o, Qwen-max) showed more stable performance in longer context windows.

\textbf{The Impact of Needle Insertion Position}. Within the configuration of the U-NIAH framework, needles in each case are inserted at intervals of 10\%, ranging from 10\% to 100\%. The different positions where the needles are inserted reflect the extent to which the LLM captures positional information in long contexts. The performance of various models with needles inserted at different positions is shown in Figure~\ref{fig:LLM_depth}. The horizontal axis represents the depth of needle insertion, while the vertical axis indicates the average score under different cases. The shade of the lines signifies different context lengths.It can be observed that across different models, there is a noticeable decline and fluctuation in performance in the mid-depth region. Particularly when the context is larger (represented by the darker blue lines), the fluctuations become more pronounced. This highlights the fact that current LLMs still face the ``Lost in the Middle" phenomenon in long contexts.

In terms of model size, smaller models (e.g., Qwen-2.5-7B and Llama-3.1-8B ) are more prone to missing information in the middle of the context and exhibit greater instability and larger fluctuations in performance. Higher-level models such as GPT4-o and Qwen-max show better stability. Furthermore, as illustrated in Figure~\ref{fig:qwen_compare}, taking the heatmaps of Qwen2.5-7B and Qwen2.5-72B with 3, 7, and 15 needles as examples, it can be seen that in regions with greater depth, when needles are inserted in the middle positions, there are more dark-colored failed cases. This indicates that the model with more parameters has better resistance to the lost in the middle effect

\begin{table*}[ht]
\centering
\caption{The performance of LLM and RAG on different cases.}
\vspace{0.4cm}
\label{tab:LLM_RAG}
\resizebox{\textwidth}{!}{%
\begin{tabular}{llcccccccccccc}
\toprule
 & \textbf{Dataset} & \multicolumn{2}{c}{\textbf{pizza-ingredients}} & \multicolumn{2}{c}{\textbf{rainbow-potion}} & \multicolumn{2}{c}{\textbf{starlight-academy-courses}} & \multicolumn{2}{c}{\textbf{starlight-teaching-methods}} & \multicolumn{2}{c}{\textbf{graduation-trials}} & \multicolumn{2}{c}{\textbf{library-revolution}} \\
 & Case & \multicolumn{2}{c}{3-Short-Needle} & \multicolumn{2}{c}{ 7-Short-Needle} & \multicolumn{2}{c}{15-Short-Needle} & \multicolumn{2}{c}{3-Long-Needle} & \multicolumn{2}{c}{7-Long-Needle} & \multicolumn{2}{c}{15-Long-Needle} \\
Model & Context / Type &  LC & RAG & LC & RAG & LC & RAG & LC & RAG & LC & RAG & LC & RAG   \\
\midrule
\multirow[t]{4}{*}{\textbf{Llama-3.1-8B}} &1-16k & 8.2±2.2 & 10.0 & 9.5±1.3 & 10.0 & 8.2±1.6 & 8.2±1.6 & 9.4±1.5 & 9.8±0.8 & 8.9±1.4 & 7.4±1.4 & 8.7±1.7 & 9.4±1.2 \\
\textbf{} & 16k-32k & 7.0±1.9 & 9.8±0.7 & 8.4±2.4 & 9.4±1.6 & 7.0±0.9 & 7.3±1.2 & 8.7±2.2 & 10.0 & 8.1±1.5 & 7.6±1.5 & 7.7±1.1 & 7.7±1.8 \\
\textbf{} & 32k-64k & 6.5±1.9 & 9.7±0.9 & 8.0±2.3 & 7.5±1.8 & 7.2±1.8 & 7.0±0.8 & 8.4±2.3 & 10.0 & 8.1±1.5 & 8.2±1.7 & 6.6±2.0 & 8.3±1.7 \\
\textbf{} & 64k-128k & 5.2±2.2 & 9.8±0.7 & 5.8±2.8 & 7.5±2.2 & 6.3±2.0 & 7.0±1.4 & 8.0±2.4 & 9.8±0.7 & 6.5±1.8 & 7.7±1.5 & 6.1±2.2 & 9.0±1.6 \\
\cline{1-14}
\multirow[t]{4}{*}{\textbf{Llama-3.3-70B}} &1-16k & 9.9±0.5 & 10.0 & 9.9±0.7 & 10.0±0.4 & 9.3±1.3 & 9.5±1.3 & 8.6±1.8 & 9.7±0.9 & 7.6±1.4 & 7.2±1.0 & 8.6±1.8 & 7.1±1.6 \\
\textbf{} & 16k-32k & 10.0 & 9.8±0.7 & 8.6±1.6 & 8.8±1.5 & 7.6±1.2 & 7.7±1.5 & 6.7±2.6 & 10.0 & 7.5±1.4 & 7.0±0.3 & 7.3±2.6 & 7.3±1.6 \\
\textbf{} & 32k-64k & 10.0 & 9.7±0.9 & 9.2±1.4 & 8.8±1.5 & 7.3±1.2 & 7.3±0.9 & 7.2±2.1 & 9.7±0.9 & 7.3±1.2 & 7.2±0.7 & 7.7±2.1 & 6.5±1.4 \\
\textbf{} & 64k-128k & 8.3±1.5 & 10.0 & 7.0±2.6 & 7.5±1.7 & 6.3±1.9 & 6.9±1.3 & 3.7±3.0 & 9.7±0.9 & 6.7±0.7 & 7.0±0.9 & 6.0±1.8 & 6.8±1.6 \\
\cline{1-14}
\multirow[t]{3}{*}{\textbf{deepseek-chat}} &1-16k & 8.0±3.3 & 10.0 & 10.0 & 10.0 & 9.7±1.0 & 9.7±0.8 & 10.0 & 10.0 & 9.6±1.0 & 8.5±1.5 & 9.5±1.1 & 9.6±1.0 \\
\textbf{} & 16k-32k & 4.9±4.5 & 9.8±0.7 & 9.6±1.3 & 9.2±1.5 & 9.3±1.2 & 7.9±1.4 & 9.8±1.1 & 10.0 & 6.7±1.8 & 8.7±1.5 & 9.6±1.0 & 9.6±1.0 \\
\textbf{} & 32k-64k & 6.5±4.3 & 9.7±0.9 & 9.0±1.6 & 7.3±1.8 & 9.7±0.9 & 7.3±0.9 & 9.8±1.1 & 10.0 & 7.3±0.9 & 9.2±1.3 & 9.2±1.2 & 9.8±0.8 \\
\cline{1-14}
\textbf{gpt-3.5-turbo} &1-16k & 9.2±1.6 & 9.7±0.8 & 7.8±2.9 & 9.1±1.4 & 8.0±1.9 & 8.6±1.5 & 9.5±1.4 & 9.9±0.6 & 7.6±1.4 & 7.5±1.2 & 6.0±1.4 & 5.8±1.0 \\
\cline{1-14}
\multirow[t]{4}{*}{\textbf{gpt-4o}} &1-16k & 10.0 & 10.0 & 9.9±0.5 & 10.0 & 9.6±1.1 & 9.8±0.8 & 10.0 & 10.0 & 10.0±0.1 & 9.0±1.1 & 9.5±1.0 & 9.2±1.2 \\
\textbf{} & 16k-32k & 10.0 & 9.8±0.7 & 10.0 & 9.4±1.2 & 9.4±1.3 & 7.1±1.2 & 9.7±1.1 & 10.0 & 9.9±0.4 & 8.8±1.3 & 9.5±0.8 & 8.8±1.0 \\
\textbf{} & 32k-64k & 10.0 & 9.7±0.9 & 10.0 & 7.5±1.6 & 8.1±1.4 & 7.2±1.7 & 10.0 & 10.0 & 9.8±0.6 & 9.0±1.0 & 9.7±0.7 & 8.8±1.7 \\
\textbf{} & 64k-128k & 9.8±0.7 & 9.7±0.9 & 9.9±0.2 & 6.9±1.8 & 8.4±1.5 & 6.9±1.4 & 10.0 & 10.0 & 9.3±1.1 & 8.9±1.3 & 9.4±0.8 & 9.3±0.9 \\
\cline{1-14}
\multirow[t]{4}{*}{\textbf{gpt-4o-mini}} &1-16k & 9.8±0.7 & 9.9±0.6 & 9.0±2.4 & 10.0 & 8.2±1.5 & 8.8±1.5 & 8.8±2.1 & 9.9±0.4 & 9.3±1.4 & 9.1±1.1 & 9.4±1.2 & 9.2±0.8 \\
\textbf{} & 16k-32k & 9.6±1.1 & 9.8±0.7 & 4.0±3.2 & 9.3±1.5 & 6.9±1.4 & 7.3±1.2 & 6.6±2.3 & 9.9±0.2 & 8.4±1.4 & 9.2±1.0 & 9.7±0.7 & 9.2±1.5 \\
\textbf{} & 32k-64k & 8.7±2.4 & 9.7±0.9 & 1.9±2.1 & 7.6±1.8 & 6.9±1.4 & 7.1±1.2 & 6.9±2.3 & 10.0 & 5.7±1.6 & 9.3±0.8 & 9.5±0.8 & 9.3±0.6 \\
\textbf{} & 64k-128k & 8.1±1.5 & 9.7±0.9 & 3.4±2.3 & 6.7±1.9 & 6.7±1.5 & 7.0±1.3 & 6.6±2.1 & 9.9±0.3 & 6.1±1.8 & 9.2±1.0 & 9.4±1.4 & 9.1±1.2 \\
\cline{1-14}
\multirow[t]{2}{*}{\textbf{qwen-max}} &1-16k & 10.0 & 10.0 & 10.0 & 10.0 & 9.7±0.8 & 9.7±1.0 & 10.0±0.2 & 9.6±0.5 & 6.9±1.4 & 7.3±0.8 & 8.0±1.1 & 9.9±0.2 \\
\textbf{} & 16k-32k & 10.0 & 9.7±0.9 & 10.0 & 9.8±0.7 & 8.1±1.5 & 7.5±1.4 & 9.8±0.7 & 9.6±0.8 & 6.4±1.3 & 7.3±0.9 & 7.6±0.9 & 10.0 \\
\cline{1-14}
\multirow[t]{4}{*}{\textbf{qwen-plus}} &1-16k & 10.0 & 10.0 & 10.0±0.4 & 10.0 & 9.3±1.2 & 9.4±1.2 & 9.9±0.3 & 9.0±1.4 & 9.7±0.6 & 9.3±0.9 & 9.5±1.0 & 8.8±0.7 \\
\textbf{} & 16k-32k & 10.0 & 9.8±0.7 & 9.8±0.7 & 9.2±1.5 & 9.0±1.4 & 7.9±1.4 & 9.9±0.3 & 8.7±1.5 & 9.5±0.9 & 9.3±0.9 & 9.6±0.6 & 8.9±0.6 \\
\textbf{} & 32k-64k & 10.0 & 9.7±0.9 & 9.7±0.9 & 7.3±1.8 & 8.4±2.2 & 7.2±1.1 & 10.0 & 8.8±1.5 & 9.3±1.1 & 8.9±1.1 & 8.9±1.3 & 8.9±0.6 \\
\textbf{} & 64k-128k & 9.2±2.3 & 10.0 & 9.6±1.1 & 7.0±2.3 & 7.0±3.1 & 7.2±1.1 & 8.3±2.8 & 8.5±1.5 & 8.9±1.3 & 9.2±0.7 & 8.0±2.3 & 9.1±0.7 \\
\cline{1-14}
\multirow[t]{4}{*}{\textbf{qwen2.5-72b}} &1-16k & 10.0±0.4 & 10.0 & 10.0 & 10.0 & 9.8±0.8 & 9.5±1.1 & 9.8±0.9 & 10.0±0.4 & 8.7±1.5 & 8.7±1.5 & 9.4±1.1 & 9.9±0.6 \\
\textbf{} & 16k-32k & 9.8±0.7 & 9.8±0.7 & 9.8±0.7 & 9.2±1.3 & 9.6±1.1 & 7.6±1.2 & 8.8±1.5 & 9.6±1.3 & 7.7±1.5 & 8.7±1.5 & 8.7±1.4 & 9.8±0.7 \\
\textbf{} & 32k-64k & 9.6±2.0 & 9.7±0.9 & 9.7±0.9 & 7.3±1.8 & 9.6±1.1 & 7.3±0.9 & 9.8±0.7 & 10.0 & 8.2±1.5 & 8.3±1.4 & 8.2±1.5 & 9.6±1.3 \\
\textbf{} & 64k-128k & 9.0±2.3 & 10.0 & 9.6±1.3 & 7.5±2.1 & 8.2±1.8 & 7.2±1.1 & 9.3±1.5 & 9.7±0.9 & 8.7±1.5 & 9.0±1.4 & 7.2±2.3 & 9.8±0.7 \\
\cline{1-14}
\multirow[t]{4}{*}{\textbf{qwen2.5-7b}} &1-16k & 8.9±1.6 & 9.8±0.7 & 7.5±1.6 & 9.8±0.8 & 7.2±1.8 & 7.5±1.8 & 8.0±2.2 & 9.9±0.6 & 7.2±1.3 & 7.3±1.4 & 9.5±1.2 & 9.8±0.7 \\
\textbf{} & 16k-32k & 7.7±1.7 & 9.8±0.7 & 7.7±2.2 & 8.8±1.8 & 7.5±1.8 & 7.2±1.1 & 7.8±2.6 & 10.0 & 6.8±1.9 & 7.2±1.1 & 8.9±1.6 & 9.6±1.3 \\
\textbf{} & 32k-64k & 7.0±2.3 & 9.3±1.5 & 7.2±2.0 & 6.9±1.8 & 5.7±1.9 & 7.2±1.2 & 6.4±2.3 & 10.0 & 6.5±1.5 & 7.5±1.1 & 7.3±2.4 & 9.8±0.5 \\
\textbf{} & 64k-128k & 5.3±2.9 & 9.8±0.7 & 6.4±2.3 & 7.2±2.2 & 5.4±1.9 & 6.8±1.7 & 5.7±2.4 & 10.0 & 5.8±2.5 & 7.2±1.2 & 6.8±1.9 & 9.3±1.3 \\
\cline{1-14}
\bottomrule
\end{tabular}
}
\end{table*}

\subsubsection{RAG vs LLM}
\noindent
The experimental results demonstrate that RAG (TopK) achieves superior performance compared to standalone LLMs in U-NIAH. Quantitative analysis reveals that RAG attains an average score of 9.04 with a standard deviation of 1.48, significantly outperforming pure LLM implementations as evidenced by an 82.58\% win-rate (defined as the proportion of experimental trials where RAG scores equal or exceed LLM scores). Complete comparative metrics are presented in Table~\ref{tab:LLM_RAG}.

\begin{figure*}[htbp]
    \centering
    \includegraphics[width=1\linewidth]{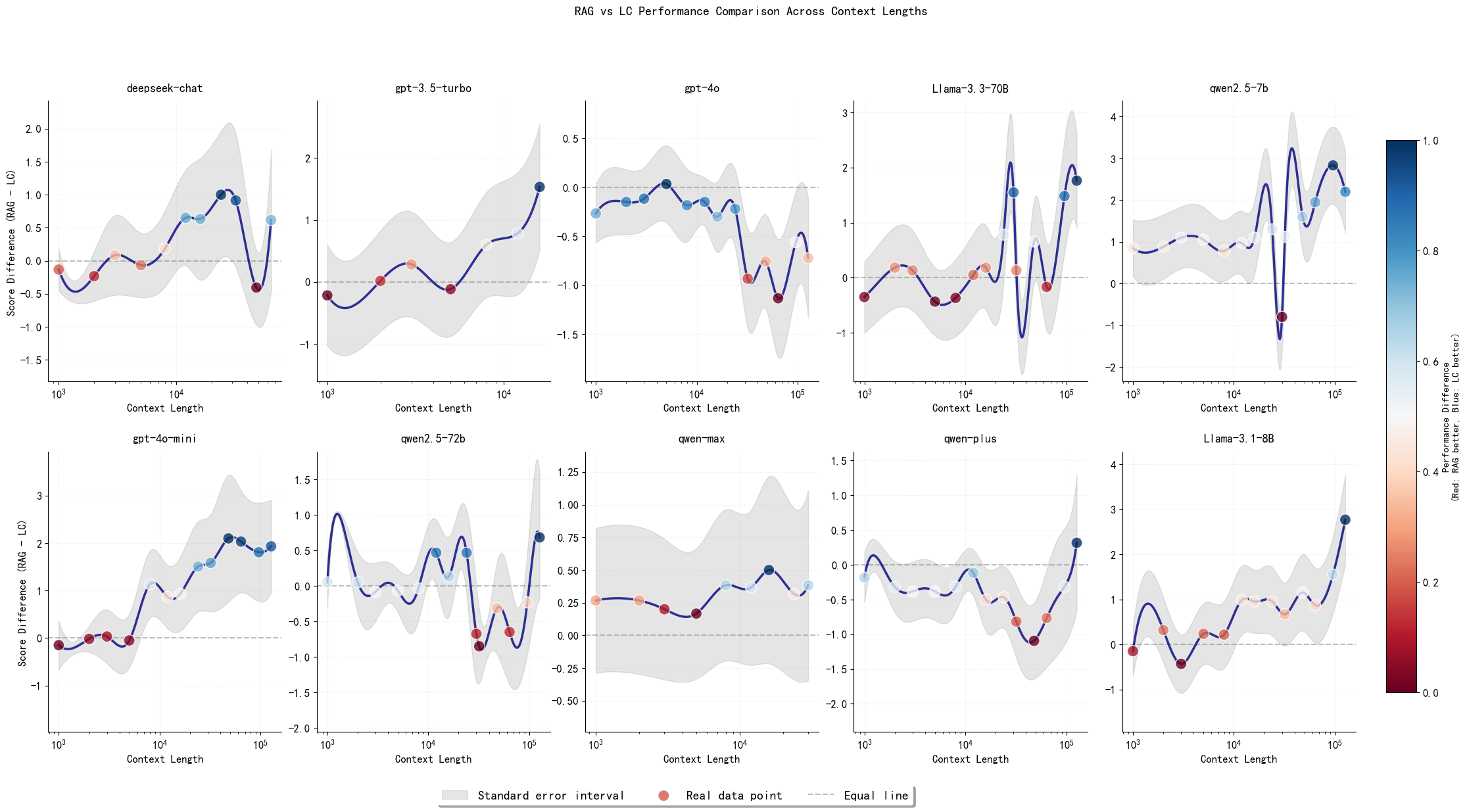}
    \caption{Performance difference between RAG and LLM in all cases.
    }
    \label{fig:LLM_RAG_diff_all}
\end{figure*}

Figure~\ref{fig:LLM_RAG_diff_all} illustrates the performance differences between RAG and LLM within the U-NIAH framework across varying context lengths. The x-axis represents the context length, while the y-axis denotes the score difference (RAG-LLM). Blue markers indicate instances where RAG outperforms LLM, whereas red markers signify superior performance by LLM. The white dashed line represents the equivalence line, where the scores of RAG and LLM are identical. The white area corresponds to the ± standard deviation error band.

It can be observed from the figure that there is a positive correlation between the expansion of context length and the improvement in RAG's performance. In the vast majority of cases, RAG outperforms LLM, with this advantage being particularly pronounced in smaller models such as Qwen2.5-7B and Llama3.1-8B. The implementation of RAG significantly enhances the ability of these models to handle long contexts.

However, it is important to note that the performance difference exhibits a dependency on model size. While RAG consistently improves long-context handling in smaller models, its effectiveness becomes less certain when applied to larger models. There are certain fluctuations, and RAG's performance does not always guarantee better results, as seen in the case of Qwen2.5-72B. This observation necessitates a more in-depth analysis. It should be emphasized that the trends presented in Figure~\ref{fig:LLM_RAG_diff_all} are derived from aggregated experiments conducted on various needle cases, which vary in difficulty. Despite their individual complexities, these cases are averaged out in the overall trend. Upon further examination of the results from these diverse cases, it is found that the improvement of RAG in smaller models is a common phenomenon. However, the performance drop of RAG in larger LLMs is not a constant occurrence.The following Figure~\ref{fig:qwen2.5_hard} provides a detailed illustration of the performance of Qwen2.5-72B in the chunk-level NIAH task. It can be observed that even in more challenging cases, RAG continues to offer benefits over standalone LLM.

\begin{figure}
    \centering
    \includegraphics[width=1\linewidth]{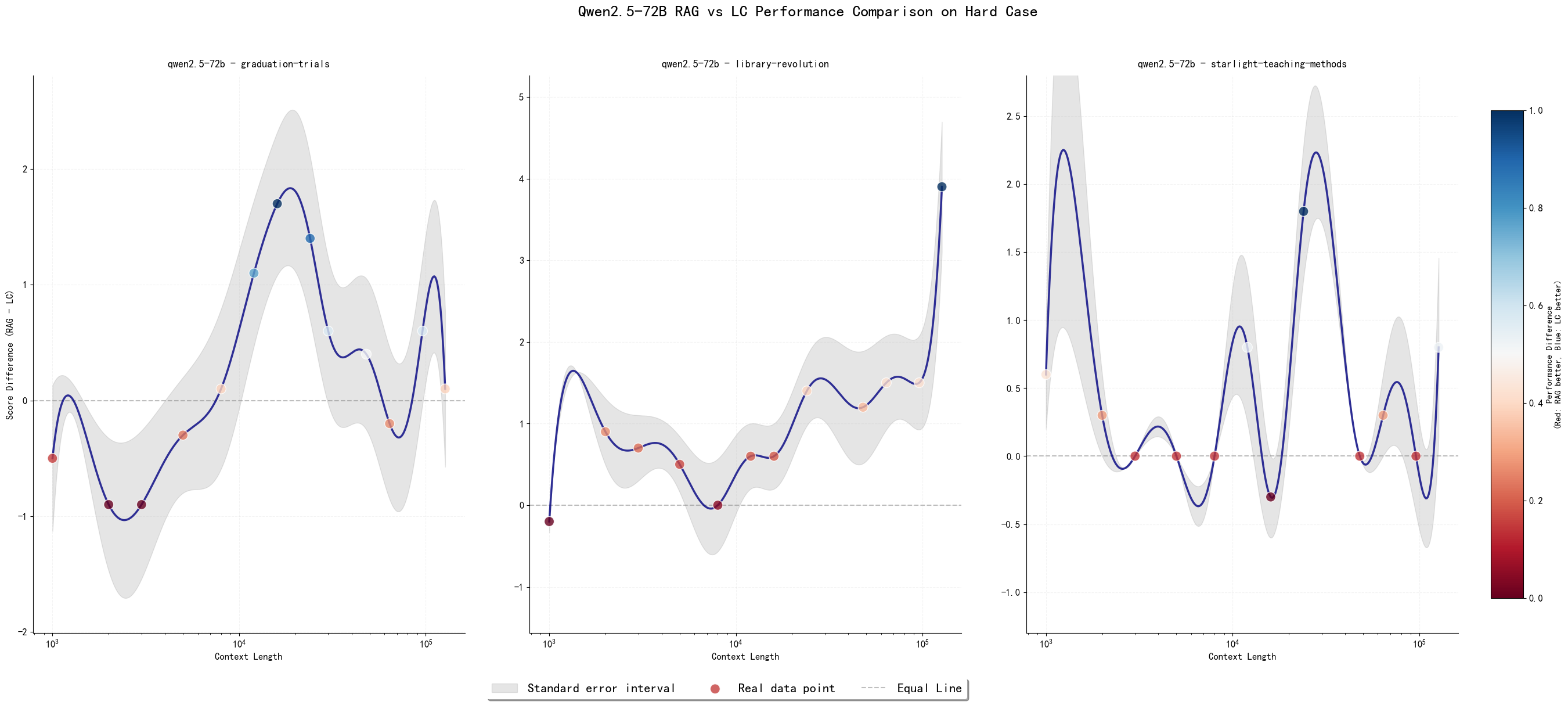}
    \caption{Performance difference between RAG and LLM of Qwen-2.5-72B on harder cases.
    }
    \label{fig:qwen2.5_hard}
\end{figure}

Overall, RAG can effectively mitigate the fluctuations of LLMs when dealing with larger Context Lengths, reduce the ``Lost in the middle" phenomenon, and enhance performance. This effect is particularly remarkable for smaller models. This indicates that when using lightweight models (such as 7B or 8B models), In-Context Learning (ICL) with long contexts should be avoided, and retrieval should be employed to reduce external irrelevant information. For more powerful models, they already exhibit satisfactory performance in simple NIAH tasks and maintain relatively stable performance even with an expanded context window. However, in more challenging scenarios, such as multi-needle and long-needle cases, they still experience performance degradation and fluctuations when the context window is long, while RAG can improve performance in such scenarios. Therefore, without considering efficiency and cost, if a more powerful LLM is used, ICL is a reasonable and direct choice in simple scenarios, and RAG is a better option in complex scenarios.

\subsection{Error Attribution and Typical Patterns in RAG}
\noindent
Previous experiments have shown that RAG is not always accurate and can still produce errors. Identifying the sources of these errors and understanding the reasons behind them is the objective of Research Question 2 (RQ2). Specifically, RQ2 aims to address the following questions: What are the causes of errors in RAG? What are the typical error patterns?

To answer RQ2, we conducted a series of experiments with enriched settings to explore the challenges faced by RAG in long-context scenarios. We introduced four additional experimental settings focusing on retrieval noise and retrieval order to investigate error attribution in long-context applications of RAG. These settings are as follows:

\textbf{Retrieval Noise Ratio}
\begin{itemize}
    \item \textbf{TopK}: Retrieving the top-K chunks, where K is a predefined parameter.
    \item \textbf{Half (Length):} Retrieving chunks until half of the given context is filled.
    \item \textbf{Full (Length)}: Retrieving chunks until the entire given context is filled.
\end{itemize}

\textbf{Retrieval Order}
\begin{itemize}
    \item \textbf{Norm}: Chunks are retrieved and arranged in ascending order of similarity, with the most relevant chunk placed farthest from the query.
    \item \textbf{Rev}: Chunks are retrieved and arranged in descending order of similarity, with the most relevant chunk placed closest to the query.
\end{itemize}

To facilitate comparative experiments and effectively measure the impact of retrieval, we selected four LLMs for our experiments, including LLaMA 3.1-8B, LLaMA 3.3-70B, Qwen 2.5-7B, and Qwen 2.5-72B. These models were tested on three Short-Needle cases to evaluate their performance in different retrieval scenarios.

By varying the retrieval noise ratio and retrieval order, we aim to uncover the underlying causes of errors in RAG when dealing with long contexts. The experimental results will provide insights into the typical error patterns and guide the optimization of RAG models for more accurate and reliable performance in long-context applications.

\begin{figure*}
    \centering
    \includegraphics[width=1\linewidth]{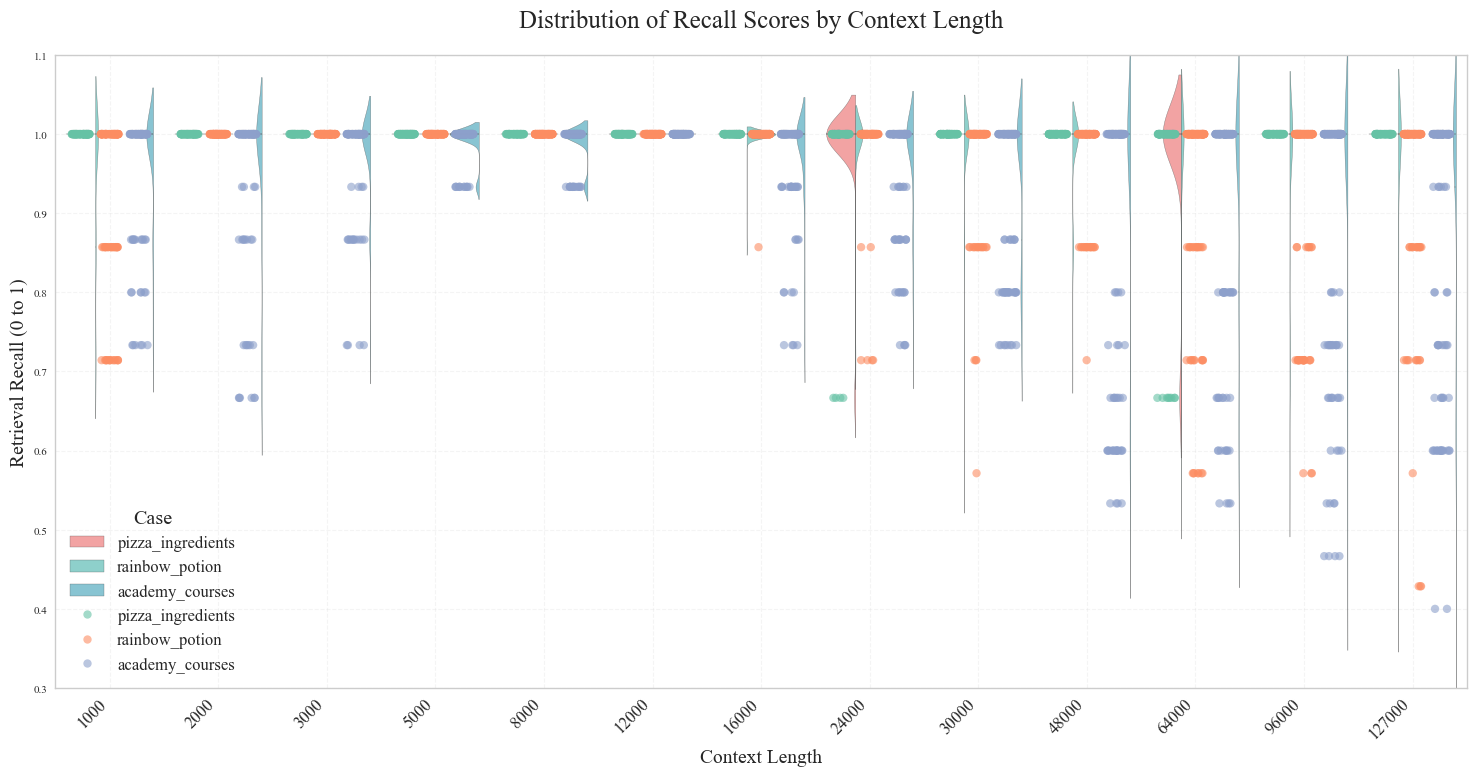}
    \caption{Retrieval Perforamnce Over Context Length}
    \label{fig:retrieval_recall}
\end{figure*}

\subsubsection{Retrieval Performance}
In RAG systems, the retrieval process determines the upper bound of the generation quality. Therefore, the recall performance of retrieval should always be the primary consideration. In this experimental group, three Short-Needle test cases (with 3, 7, and 15 needles, respectively) achieved overall retrieval completeness rates of 99.03\%, 90.78\%, and 79.59\%, respectively. These results were obtained using three different retrieval scopes: TopK, Full Length, and Half Length. Retrieval failures were mainly concentrated in the TopK setting, while the Half Length and Full Length settings provided more redundancy and thus better robustness. 

Figure~\ref{fig:retrieval_recall} illustrates the retrieval recall rates of the three cases under different context lengths. It is evident that as the number of needles increases, the difficulty of retrieval also rises. Additionally, the longer the context, the greater the fluctuation in retrieval performance. This is because longer contexts contain more distracting chunks, which increase the complexity of the retrieval task. Notably, in the 15-needle test with a 127K context, the lowest retrieval recall rate was only 40\%. This highlights the significant challenge of retrieving relevant information from large contexts, especially when the number of needles is high.

\subsubsection{Error Type under Perfect Retrieval}
\noindent
Based on the statistical analysis of the generated results, we have categorized the error types into five distinct categories: 
(1) \textbf{Only Missing Elements}; (2) \textbf{Only Wrong Elements} (LLM give nonexistent Needle); (3) \textbf{Both Types of Errors} (Missing and Wrong); (4) \textbf{Self-Doubt with Complete Answer} (correctly identifying all relevant elements but refusing to answer or give a negative answer); (5) \textbf{Self-Doubt with Incomplete Answer}; (6) \textbf{Others} (such as nonsensical responses).

\begin{figure}[htbp]
    \centering
    \includegraphics[width=1\linewidth]{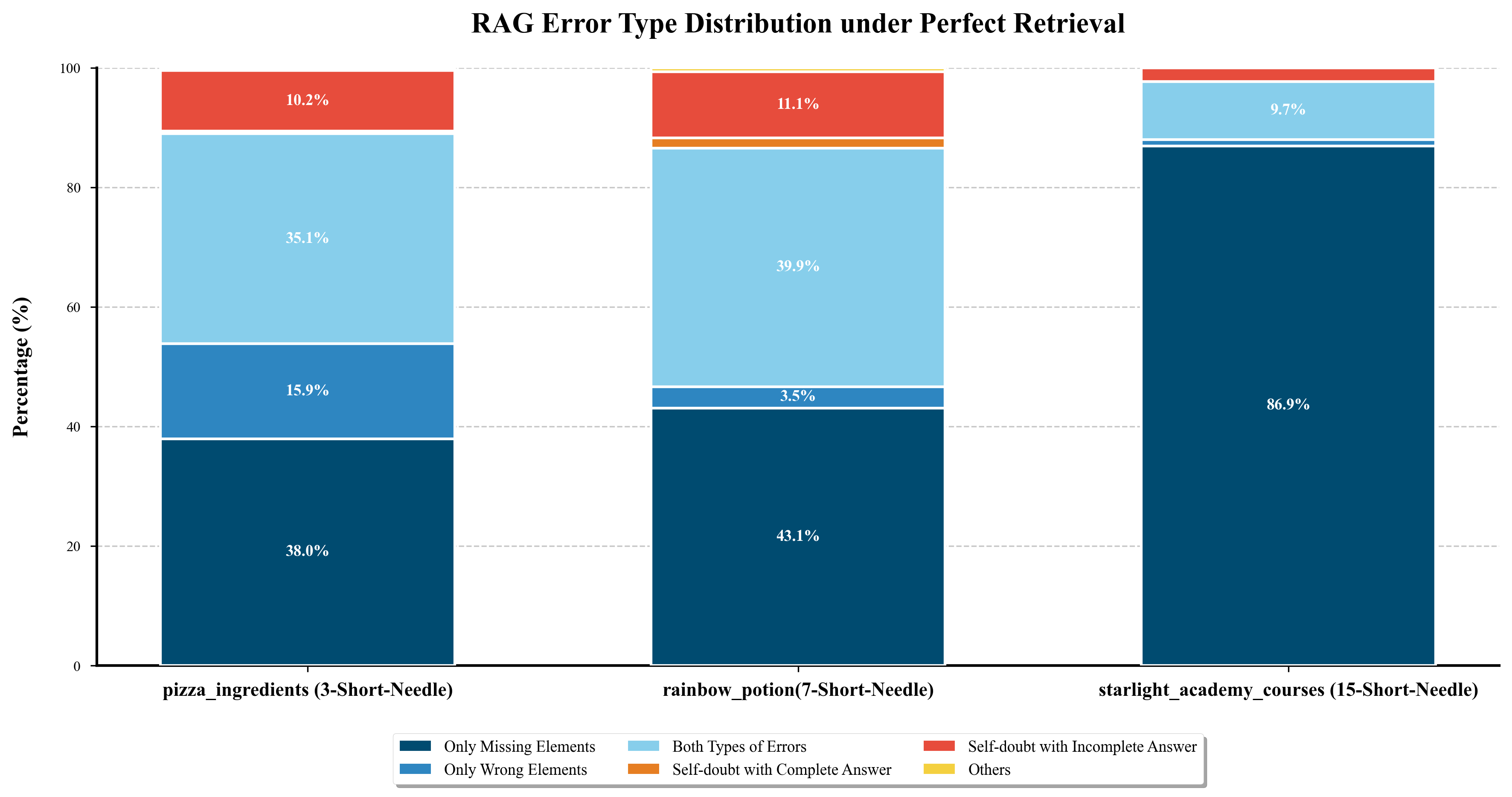}
    \caption{RAG Error Type Distribution under Perfect Retrieval}
    \label{fig:error_type}
\end{figure}

\begin{figure}[htbp]
    \centering
    \includegraphics[width=1\linewidth]{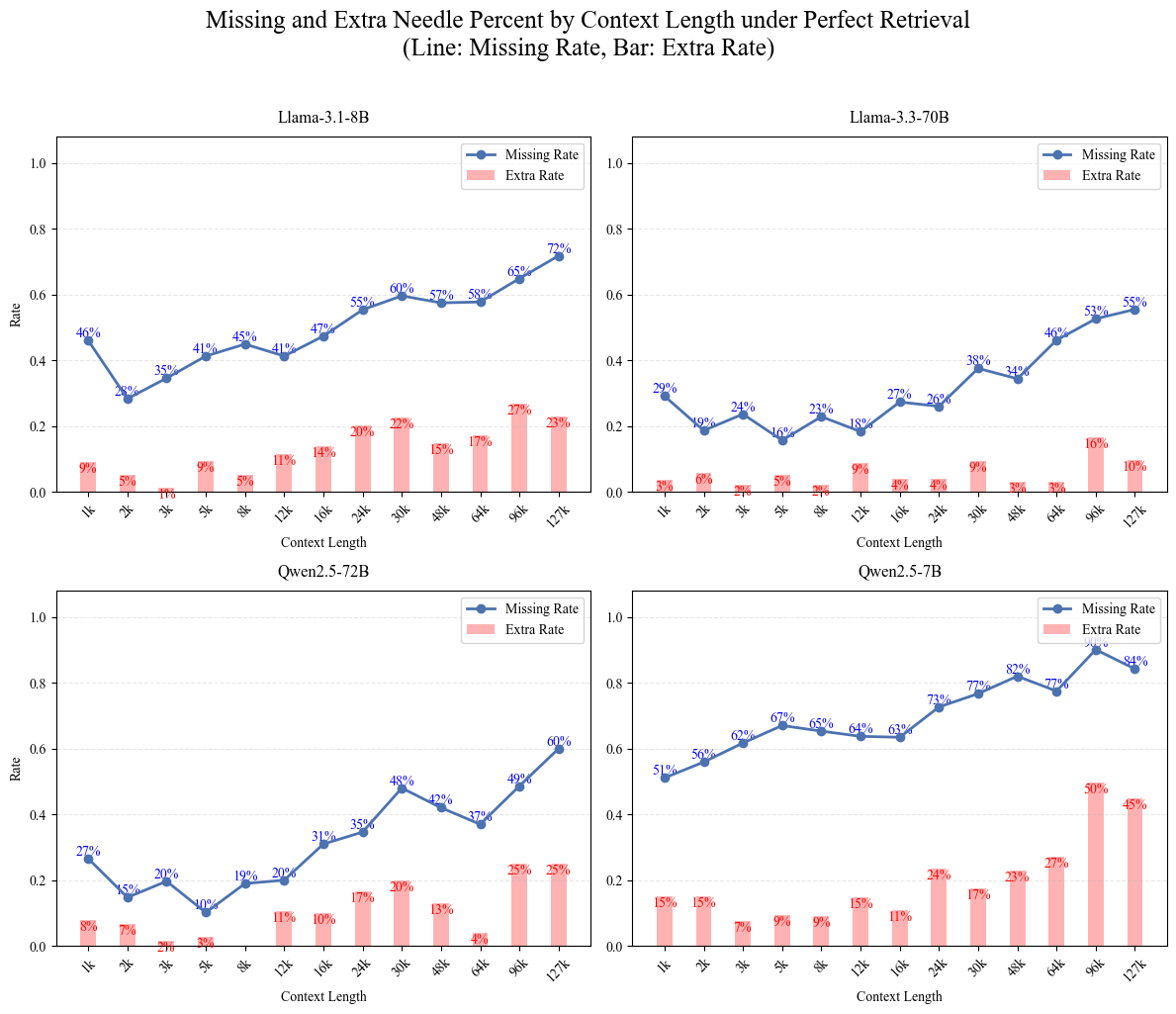}
    \caption{Missing and Extra Needle Percent by Context}
    \label{fig:error_context}
\end{figure}

Figure~\ref{fig:error_type} illustrates the error distribution of RAG across three cases, excluding the scenarios where retrieval is incomplete. It is evident that even with a complete retrieval phase, omission remains the most significant error type. Moreover, as the number of relevant elements (Needles) increases, the severity of omission errors becomes more pronounced. In the 15-Needle case (starlight\_academy\_courses), approximately 86.9\% of the errors are attributed solely to omission. Furthermore, in cases with fewer Needles, hallucination errors and mixed errors (both omission and hallucination) still account for a considerable proportion. For instance, in the 3-Needle case, 51\% (35.1\% + 15.9\%) of the errors involve hallucination. This indicates that when the amount of key information is limited and noise is abundant, LLMs tends to rely on its background knowledge to generate responses, thereby producing hallucinations. In addition to the basic error types of omission and hallucination, the LLM exhibited behaviors of self-doubt and refusal to answer during the experiments. In many instances, LLM did not fail to identify the relevant Needles but rather lacked confidence in its answers, resulting in self-doubt. 

One scenario is when the LLM considers the overall text to be irrelevant , leading to self-doubt (`\textit{Needle1, Needle2..., but they are not related to the text}'). Another situation is when the LLM's background knowledge interferes with its responses. For example, the model correctly listed all the elements in the standard answer (`\textit{Prosciutto, goat cheese, and figs}'), thus achieving completeness in terms of elements. However, the model's response included a negative evaluation of these ingredients (`\textit{but it is actually a mistake as they are not typically used in pizza making}'), demonstrating self-negation. This is precisely why we fully fabricate a fictional world in the U-NIAH framework to avoid interference from external knowledge.

Another manifestation of self-negation is when the model correctly lists all seven elements in the standard answer but begins its response with a degree of self-negation (`\textit{The seven colors that make up the magical rainbow are not explicitly mentioned...}'). This indicates a lack of confidence in its own response. Therefore, even though the answer is complete and accurate, the model's display of self-doubt is noteworthy.

Figure~\ref{fig:error_context} illustrates the variation of omission and hallucination errors with respect to context length under three different settings: TopK, Half-Length, and Full-Length. The figure clearly demonstrates that as the context length increases, both omission and hallucination errors also tend to rise. This trend is primarily attributed to the introduction of additional noise in the Half-length and Full-length settings. This finding indicates that retrieving more content does not necessarily guarantee improved performance. In fact, it may lead to increased errors due to the higher likelihood of irrelevant information being included.

Moreover, smaller models are found to be more susceptible to omission errors and are more prone to generating hallucinations. This suggests that limited model capacity may hinder the effective identification and filtering of relevant information, thereby exacerbating the issues of both missing critical elements and introducing non-existent ones.

\begin{figure}[ht]
    \centering
    \includegraphics[width=1\linewidth]{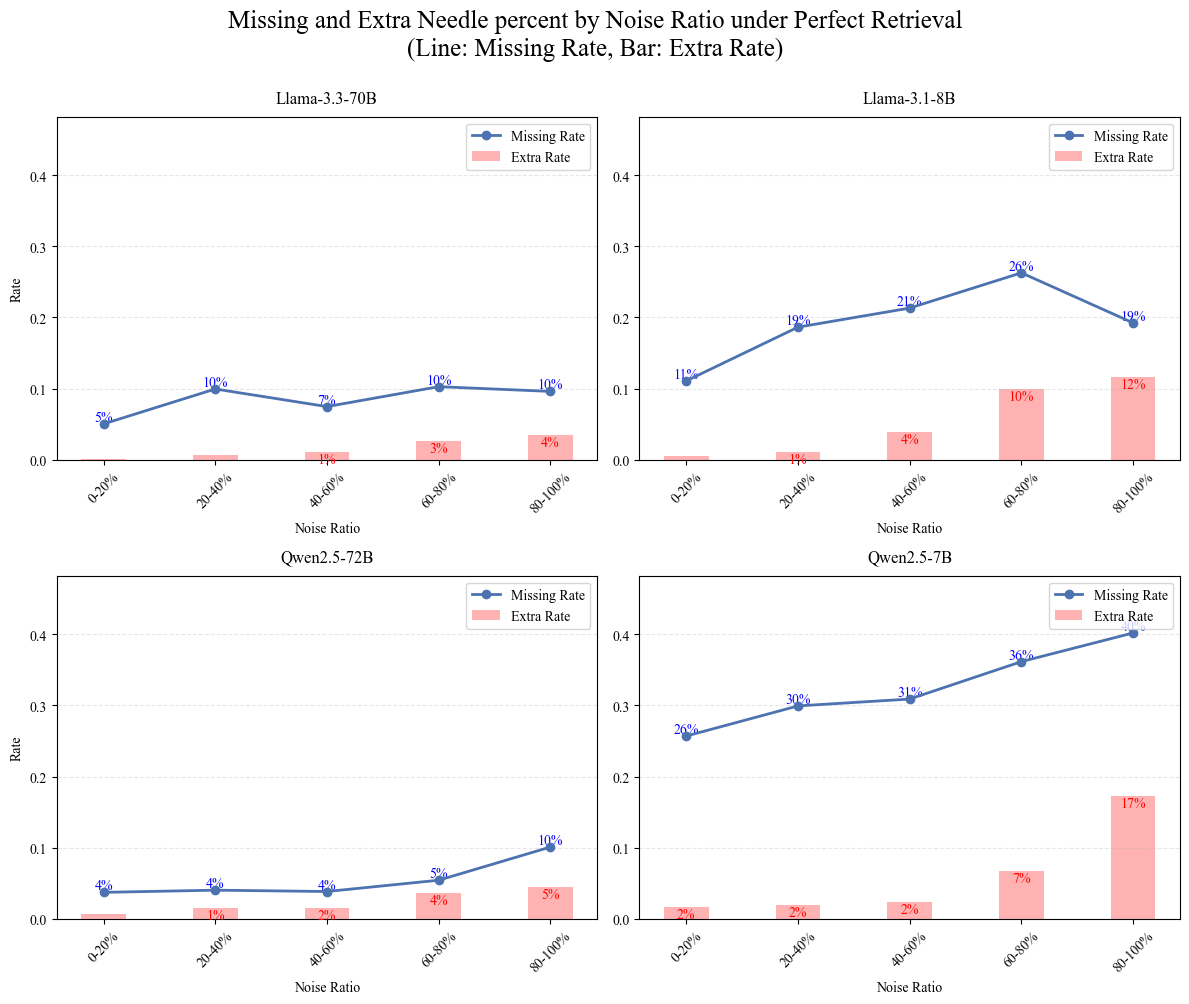}
    \caption{The Effect of Noise Ratio}
    \label{fig:noise_ratio}
\end{figure}

\subsubsection{Impact of RAG Context Organization}
\noindent
We further investigate the impact of context organization methods in RAG on the final performance. Table~\ref{tab:RAG_Context_org} summarizes the scores under different context settings, including three retrieval scopes (Topk, Half-Length, and Full-Length) and two chunks ordering schemes (Norm and Rev).

\textbf{Noise Ratio.} From the experiments, it is evident that retrieving more information does not always guarantee better results. In the three-needle case (pizza\_ingredients), the TopK setting achieved the best performance across all context stages. In the multi-needle cases with seven and fifteen needles (rainbow\_potion and academy\_courses), the Half-length setting performed the best, followed by Topk, and then Full-Length. Considering the difficulty of the three cases and their corresponding retrieval effects, the improvement of half-length over Topk is primarily due to the broader retrieval scope, which ensures a higher likelihood of including more relevant elements (Needles) in the context. In contrast, the Full-length setting introduces more noise, leading to a decline in performance. Further analysis of different models under Half-length and Full-length settings reveals that smaller models exhibit a more significant performance drop when transitioning from half to full length. This aligns with our findings in the pure LLM setting, where smaller models still demonstrate weaker noise resistance in the RAG context. Additionally, in each experimental setting, we observe a decline in performance as the context size increases. This indicates that a larger context not only poses challenges for retrieval (in the Topk setting), where a larger context increases the likelihood of strong distractors, but also presents difficulties in effectively utilizing the retrieved chunks.

To further quantify the impact of noise on experimental results, we designed the Noise Ratio metric, which measures the proportion of chunks in the context that do not contain the correct Needle relative to the total number of chunks used in the context. As shown in Figure~\ref{fig:noise_ratio}, overall, the proportions of omission and hallucination errors increase with the rise in noise levels. Larger models are less affected by noise in terms of omission errors but experience an increase in hallucination errors. In contrast, smaller models are more significantly impacted by noise. For instance, in the llama-3.1-8B and Qwen2.5-7B, both omission and hallucination errors increase substantially as the noise ratio rises.

\begin{table}[htbp]
\centering
\caption{Performance of Different Context Organization}
\label{tab:RAG_Context_org}
\vspace{0.4cm}
\resizebox{0.5\textwidth}{!}{
\begin{tabular}{llrrr|rr}
\toprule
\textbf{Case} & \textbf{Context} & \textbf{Topk} & \textbf{Half} & \textbf{Full} & \textbf{Norm} & \textbf{Rev} \\
\midrule
\multirow[t]{4}{*}{\textbf{pizza\_ingredients}} & 0-16K & 9.95 & 9.23 & 9.67 & 9.65 & 9.32 \\
\textbf{} & 16K-32K & 9.77 & 8.17 & 9.14 & 9.22 & 8.10 \\
\textbf{} & 32K-64K & 10.00 & 7.96 & 9.42 & 9.09 & 8.44 \\
\textbf{} & 64K-128K & 9.49 & 7.93 & 8.39 & 8.75 & 8.06 \\
\cline{1-7}
\multirow[t]{4}{*}{\textbf{rainbow\_potion}} & 0-16K & 9.43 & 8.56 & 8.67 & 9.12 & 8.40 \\
\textbf{} & 16K-32K & 8.18 & 8.45 & 8.10 & 8.41 & 8.15 \\
\textbf{} & 32K-64K & 7.56 & 8.65 & 7.71 & 8.05 & 8.11 \\
\textbf{} & 64K-128K & 7.20 & 8.06 & 7.07 & 7.49 & 7.59 \\
\cline{1-7}
\multirow[t]{4}{*}{\textbf{academy\_courses}} & 0-16K & 8.40 & 7.61 & 7.23 & 8.17 & 7.41 \\
\textbf{} & 16K-32K & 7.21 & 7.53 & 6.67 & 7.46 & 6.92 \\
\textbf{} & 32K-64K & 6.83 & 7.00 & 6.03 & 6.74 & 6.60 \\
\textbf{} & 64K-128K & 6.71 & 6.86 & 5.25 & 6.37 & 6.29 \\
\cline{1-7}
\bottomrule
\end{tabular}
}
\end{table}

\textbf{Order of Chunks.} In terms of the organization of retrieved documents, sequential ordering (Norm) demonstrated significantly better performance. This indicates that the arrangement of retrieved documents indeed affects the LLM's ability to understand the content. This suggests that when dealing with longer contexts, LLMs tend to place higher implicit emphasis on the beginning of the context. Sequential ordering places highly relevant documents at the forefront, effectively occupying the model's ``prime memory zone". From another perspective, the presence of a large amount of irrelevant information at the beginning of the document can interfere with the LLM's understanding of the overall document content. This is contrary to the style of many pretraining and Supervised Fine-Tuning (SFT) corpora. Therefore, even though key documents are closer to the query in reverse ordering, this arrangement disrupts the LLM's ability to comprehend critical information within the longer context. This phenomenon is also influenced by the model's parameter size. After averaging the scores across all context lengths, the performance degradation caused by reverse ordering compared to sequential ordering for four different models is as follows:
\textbf{Llama3.1-8B (-16.89\%)};
Llama3.3-70B (-3.90\%);
Qwen2.5-72B (-0.37\%);
\textbf{Qwen2.5-7B (-10.75\%)};
It is evident that smaller models are more susceptible to the negative impact of reverse ordering.

\begin{table*}[h]  
\centering  
\caption{Summary of Error Patterns in RAG under U-NIAHFramework}  
\label{tab:error_patterns}  
\vspace{0.5cm}
\resizebox{\textwidth}{!}{%  
\begin{tabular}{  
  @{}  
  >{\raggedright\arraybackslash}m{0.15\textwidth} % Modified column types  
  >{\raggedright\arraybackslash}m{0.2\textwidth}  
  >{\raggedright\arraybackslash}m{0.08\textwidth}  
  >{\raggedright\arraybackslash}m{0.08\textwidth}  
  >{\raggedright\arraybackslash}m{0.12\textwidth}  
  >{\raggedright\arraybackslash}m{0.12\textwidth}  
  >{\raggedright\arraybackslash}m{0.2\textwidth}  
  @{}  
}  
\toprule  
\textbf{Error Pattern} & \textbf{Condition} &\textbf{ Behavior} & \textbf{Effect} & \textbf{LC Impact} &\textbf{ Model Size} &\textbf{ Cause} \\
\midrule  
Retrieval Bottleneck & Recall $<$ 1 & Missing Element & 10.18\% & Increase with longer LC & Independent & Limits the upper bound of generation \\
\addlinespace  
Noise Distraction & Noise Ratio $>$ 44\% & Missing Element & +33.53\% & Increase with longer LC & Smaller more affected & High noise interferes with LLM's information capture \\
\addlinespace  
Noise Critical & Long Context + Noise Ratio $>$ 97\% & Extra Element & +355.82\% & Occurs in long contexts & Smaller more affected & Hallucination surges with high noise at critical context length \\
\addlinespace  
Position-sensitive & Context Length $<$ 16K + Reverse & Missing Element & +59.64\% & Pronounced in small / medium contexts  & Smaller more affected & Irrelevant info at the beginning interferes with understanding \\
\addlinespace  
Self-consistency & Small Model + Context $>$ 16K + Reverse & Self-doubt & 11.30\% & Significant in longer LC & Smaller models 377\% more  & Incoherent text in long contexts causes self-consistency errors \\
\bottomrule  
\end{tabular}%  
}  

\end{table*}  

\subsubsection{Typical Error Patterns}
\noindent
In summary, based on the aforementioned analysis, this paper further summarizes the error pattern as shown in Table~\ref{tab:error_patterns}. Initially, the limitation of the retrieval bottleneck results in the primary outcome of missed needles. Across all experimental tests, a total of 10.18\%  of cases were attributed to incomplete retrieval, which is unrelated to the generative model.

Noise distraction errors refer to the retrieval of excessive irrelevant chunks, which in turn interferes with the LLM's ability to capture relevant information. Utilizing DecisionTreeRegressor to automatically detect thresholds and employing statistical tests to ensure significant differences between groups, it was concluded that when the Noise Ratio exceeds 44\%, the proportion of missed elements increases by 33.53\%. Moreover, the validation of context length intervals revealed that this phenomenon is positively correlated with Context Length.

The noise critical pattern occurs under conditions of large Context sizes (approaching the maximum Context window) combined with extremely high noise ratios (Noise Ratio $>$ 97\%), where the model's hallucination is significantly amplified, leading to a 355.82\% increase in the probability of generating false elements. This situation is predominantly observed under the RAG Half and Full-Length setting. The critical value is also calculated by DecisionTreeRegressor and verified through statistical tests. This pattern primarily occurs in larger contexts and has a more pronounced impact on smaller models.

Position-sensitive errors pertain to the retrieval of chunks in reverse order. Group analysis indicates that within a 16K context window, the reverse order leads to a significantly higher rate of element omission (+59.4\%), with smaller models being more affected. We attribute this to the fact that smaller models tend to focus more on the initial information, and the irrelevant information at the beginning interferes with the model's understanding of the context.

Self-consistency errors occur when, despite capturing the correct elements, the LLM generates incorrect answers due to self-doubt. This phenomenon is more pronounced in smaller models, which have a 377\% higher probability of self-doubt compared to larger models. Moreover, the occurrence of this error is highest, reaching 11.30\%, when the Context exceeds 16K and the RAG chunk organization is in reverse order.

Overall, based on the identified error patterns, several actionable insights emerge to enhance the practical deployment of RAG systems. First, optimizing the retrieval phase is critical to minimize the missing information. Developers should prioritize noise control to prevent irrelevant content from overwhelming the LLM’s attention, particularly in longer contexts where interference escalates. For scenarios involving large context windows, strict safeguards against noise ratios are essential to avoid amplified hallucination risks, especially in smaller models. Positional sensitivity highlights the importance of maintaining a sequential information order during chunk organization, with more relevant content placed first. Reversed sequences disproportionately degrade performance in smaller models, likely due to their bias toward early context. Practitioners should balance the number of retrieved chunks with noise tolerance, favoring moderate window sizes and sequential chunk arrangements while leveraging larger models where feasible to mitigate multiple error patterns simultaneously. These findings collectively advocate for context-aware RAG tuning, dynamic noise filtering, and model-capability-aligned design choices to improve RAG reliability across diverse applications.

\subsection{Harder case: Needle in the Needle}
\noindent
To address RQ3: What are the key factors that RAG needs to optimize in the LC setting to address more complex scenarios? we designed a challenging needle-in-the-needle scenario that unifies two critical interference factors in RAG : hard negative samples and information fragmentation across split chunks. This synthetic testbed examines scenarios where critical information becomes vulnerable to both segmentation artifacts during chunking and semantically similar distractors in the retrieval stage.

The experimental configuration embeds smaller critical information fragments (``short-needles") within larger contextual segments (``long-needles"), which are then distributed across background text. Each short-needle containing Starlight Academy's first-year course information as long-needles was intentionally designed to exceed standard chunk sizes, ensuring mandatory segmentation across multiple chunks. The short-needles (15 target courses) were non-uniformly distributed among long-needles, creating varying difficulty levels - some long-needles contained multiple short-needles while others contained none. Background noise included semantically related distractors from Starlight Academy's upper-year curricula (Grades 2-4), competing institution courses (Lunaris Academy),and we also use domain-irrelevant content from Paul Graham Essays as a comparsion.

This dual-challenge design simultaneously tests two critical RAG capabilities: 1) Robustness against information fragmentation through mandatory chunk splits and 2) Discrimination power against hard negative samples. We further extended the evaluation framework by incorporating emerging reasoning LLMs (e.g.,O3-mini and Deepseek-R1) that employ inference-time scaling to enhance complex reasoning. This extension allows preliminary investigation of potential synergies between enhanced reasoning LLMs and RAG, particularly for NIAH tasks that requiring multi-chunk evidence synthesis. The comparative analysis aims to identify complementary strengths between RAG and most powerful LLMs in handling more difficult challenges.

\begin{table*}[ht]
\centering
\caption{Performance of RAG and LLM on Needle-in-Needle}
\vspace{0.4cm}
\label{tab:RQ3}
\resizebox{\textwidth}{!}{%
\begin{tabular}{l lcccccccccc}
\toprule
 &\textbf{ Model} & \multicolumn{2}{r}{\textbf{deepseek-chat}} & \multicolumn{2}{r}{\textbf{deepseek-reasoner}} & \multicolumn{2}{r}{\textbf{gpt-4o-mini}} & \multicolumn{2}{r}{\textbf{o1-mini}} & \multicolumn{2}{r}{\textbf{o3-mini}} \\
 % & type & LC & RAG & LC & RAG & LC & RAG & LC & RAG & LC & RAG \\
Dataset & Context / Type & LC & RAG & LC & RAG & LC & RAG & LC & RAG & LC & RAG \\
\midrule
\multirow[t]{4}{*}{\textbf{PaulGrahamEssay-needle-in-needle}} & 1-16k & 8.0±1.1 & 7.5±0.9 & 9.0±0.6 & 8.0±1.0 & 7.0±0.8 & 7.3±0.9 & 7.7±1.2 & 7.0±1.2 & 7.8±2.3 & 9.1±1.0 \\
\textbf{} & 16k-32k & 7.7±1.1 & 7.7±0.9 & 8.7±1.2 & 8.2±1.0 & 6.7±1.0 & 7.3±0.8 & 6.8±1.0 & 6.8±0.9 & 7.8±1.7 & 9.1±1.1 \\
\textbf{} & 32k-64k & 7.4±0.8 & 7.9±1.2 & 9.3±0.5 & 8.6±1.3 & 6.2±1.0 & 7.8±1.1 & 6.3±1.0 & 7.1±1.3 & 8.0±1.6 & 9.2±1.2 \\
\textbf{} & 64k-128k & -- & -- & -- & -- & 6.3±1.2 & 7.8±1.1 & 5.8±1.0 & 7.0±1.4 & 7.6±1.4 & 8.8±1.2 \\
\cline{1-12}
\multirow[t]{4}{*}{\textbf{StarlightAcademy2-needle-in-needle}} & 1-16k & 8.5±1.0 & 7.7±1.4 & 9.7±0.5 & 6.7±1.2 & 7.1±0.4 & 7.9±1.2 & 9.4±0.7 & 8.1±1.4 & 9.6±0.6 & 9.1±1.2 \\
\textbf{} & 16k-32k & 7.9±1.1 & 8.0±1.9 & 9.3±0.7 & 5.7±2.7 & 6.9±0.8 & 8.4±1.2 & 8.9±1.2 & 9.9±0.3 & 8.8±0.9 & 9.4±1.9 \\
\textbf{} & 32k-64k & 8.0±0.9 & 7.8±1.3 & 9.3±1.1 & 5.8±3.4 & 6.8±0.6 & 9.4±0.7 & 7.8±0.9 & 9.1±1.7 & 8.8±1.0 & 9.3±2.2 \\
\textbf{} & 64k-128k & -- & -- & -- & -- & 7.0±0.5 & 9.6±0.7 & 7.6±2.0 & 9.3±1.4 & 9.0±0.6 & 9.5±1.7 \\
\cline{1-12}
\bottomrule
\end{tabular}
}
\end{table*}

\subsubsection{Overall Performance}
\noindent
The experimental results presented in Table~\ref{tab:RQ3} reveal significant insights into the performance dynamics between RAG and LLM approaches in specifically designed challenging Needle-in-Needle scenarios. Contrary to their demonstrated superiority in conventional tasks, RAG systems exhibit comparable effectiveness to standalone LLMs in U-NIAH framework, with mean scores of 8.04 and 8.01 respectively. 

These findings validate our initial hypothesis that strategically put critical information (Short-Needle) within extended contexts (Long-Needle) containing strong distractors, coupled with intentional document fragmentation across multiple chunks, substantially diminishes RAG's conventional advantages. This phenomenon can be attributed to the inherent challenges in effective knowledge retrieval under such adversarial conditions.

Notably, despite the overall parity in performance metrics, RAG demonstrates superior robustness by achieving equivalent or better results than LLMs in about 70\% of test cases. This resilience advantage becomes particularly evident when analyzing performance across varying context lengths. Segmentation analysis reveals a progressive improvement in RAG's win rate relative to LLMs across increasing context ranges: 57.4\% in [1-16k], 77\% in [16-32k], 86.7\% in [32-64k], and 92.7\% in [64-128k] token intervals. This pattern suggests that RAG's architectural advantages become more pronounced as context complexity escalates, potentially due to its reduced susceptibility to context-length-induced performance degradation.

The investigation into background noise effects yields additional critical insights. RAG maintains a 74.7\% win rate when confronted with unrelated informational noise (PaulGrahamEssay scenarios), compared to 65\% in environments containing relevant contextual noise (StarlightAcademy2). This 9.7\% performance differential underscores the critical impact of semantic relevance in retrieval interference, indicating that competing information sharing conceptual relationships with target content (e.g, course from other Grade) poses greater challenges to RAG's retrieval mechanisms than generic noise. These findings collectively suggest that while RAG's absolute performance advantages may diminish under specifically engineered adverse conditions, its architectural design preserves fundamental robustness benefits compared to direct LLM application, particularly in complex operational environments.

\begin{figure*}[htbp]
    \centering
    \includegraphics[width=1\linewidth]{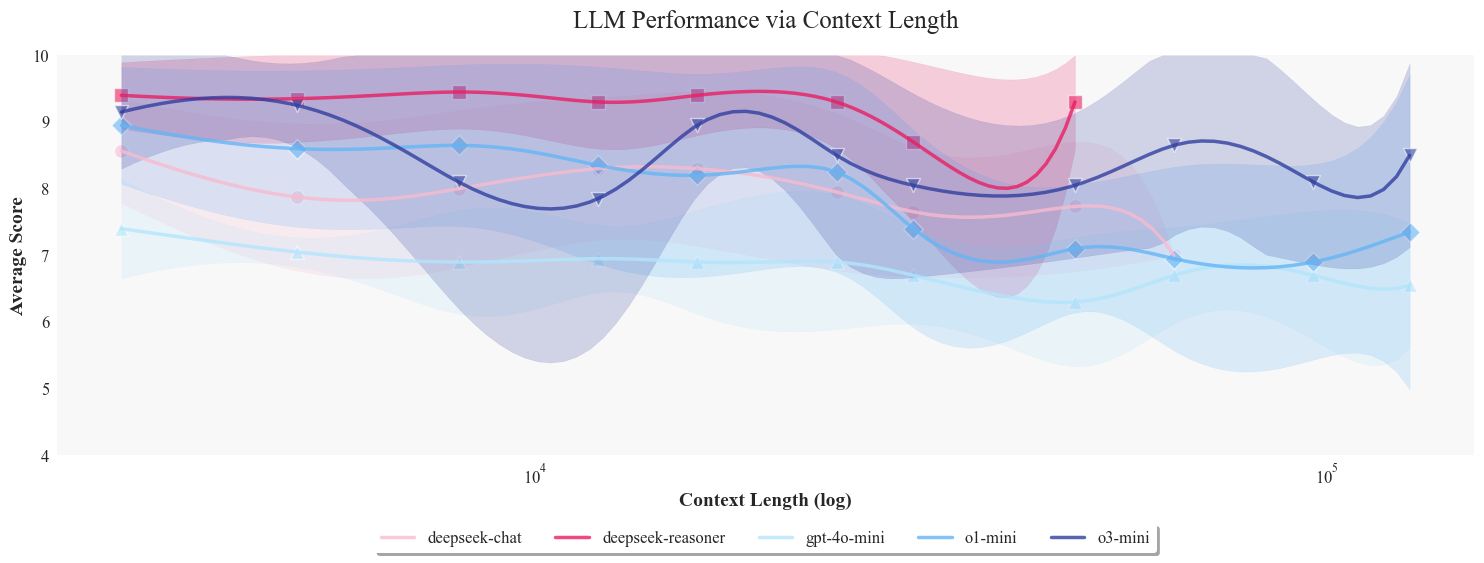}
    \caption{LLM Performance on Needle-in-Needle Case}
    \label{fig:LLM_performance_RQ3}
\end{figure*}

\subsubsection{Reasoning LLMs}
\noindent
This section investigates the performance of deliberate reasoning (``slow-thinking") LLMs in complex scenarios and their synergistic compatibility with RAG systems. Figure~\ref{fig:LLM_performance_RQ3} illustrates the comparative performance of various LLM configurations across increasing context lengths in NIAH tasks, with chromatic coding distinguishing model families,darker hues denote deliberate reasoning models. Specifically, deep red represents DeepSeek-Reasoner (R1), while navy blue corresponds to OpenAI's O3-mini.  

The analysis reveals that deliberate reasoning LLMs demonstrate superior efficacy in processing extended contexts. DeepSeek-Reasoner achieves the highest performance, followed by O3-mini, with both significantly outperforming baseline models. Notably, within the OpenAI mini-series hierarchy (GPT-4o-mini, O1-mini, O3-mini), we observe a progressive performance enhancement correlating with architectural complexity. This gradient suggests that slow-thinking mechanisms substantially augment LLMs' capacity to identify and synthesize critical information from lengthy, information-dense passages. 

These findings empirically demonstrate that thinking processes that incorporate explicit reasoning loops effectively mitigate cognitive overload in long-context scenarios. Furthermore, the consistent performance hierarchy across model scales implies that reasoning capabilities operate orthogonally to parameter count, presenting opportunities for optimized deployment of smaller, specialized models in latency-sensitive applications requiring long-context comprehension.

\begin{figure*}
    \centering
    \includegraphics[width=1\linewidth]{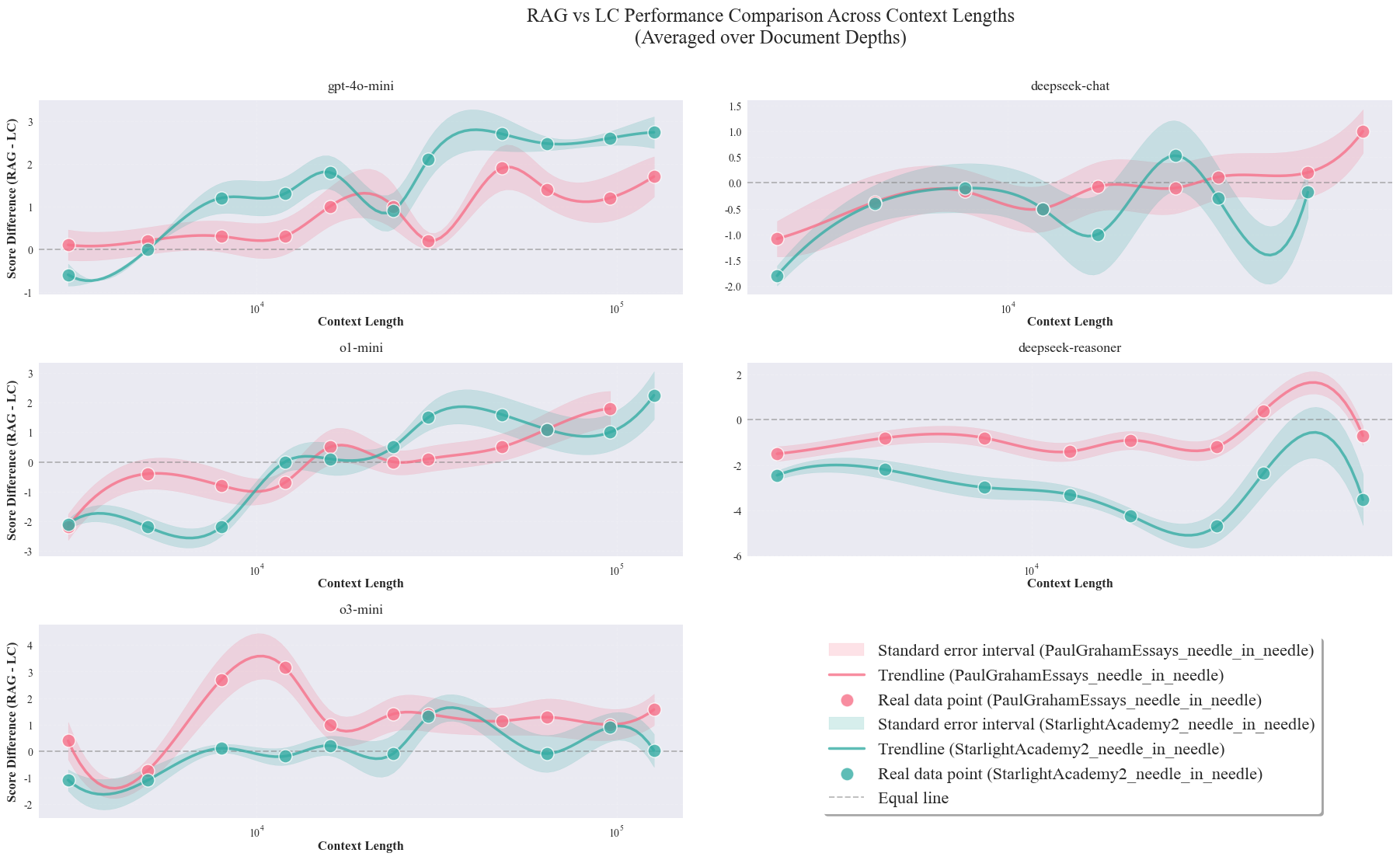}
    \caption{RAG and LLM Performance Comparsion in Needle-in-Needle}
    \label{fig:RAG_LLM_RQ3}
\end{figure*}

Figure~\ref{fig:RAG_LLM_RQ3}illustrates the performance differentials between RAG and LC across multiple datasets, with each data point representing the score difference at identical context lengths, aggregated across various needle insertion depths. The visualization employs green markers to denote experiments using relevant background corpus (StarlightAcademy2) containing needle-related content, while red markers indicate usage of irrelevant corpus (PaulGrahamEssays) as filler text. The left panel displays results from OpenAI's mini-series models, contrasted with DeepSeek series models on the right.

Experimental findings demonstrate that RAG generally enhances NIAH task performance across most models (excluding O3-mini and DeepSeek-Reasoner), with performance gains showing positive correlation to increasing context lengths. Notably, more capable reasoning models exhibit diminished returns from RAG augmentation, particularly evident in DeepSeek-Reasoner's performance where RAG implementation frequently underperforms baseline LLM results. It further reveals that advanced reasoning models (O3-mini and DeepSeek-Reasoner) display heightened sensitivity to background noise.

This observation suggests that slow thinking exhibit weaker interference resistance against hard negative examples, which RAG inadvertently amplifies through concentrated retrieval of disturbing evidence, and excessive deliberative processing may introduce instability in RAG workflows. These findings collectively indicate that while reasoning models demonstrate superior performance in in-context learning scenarios, they present implementation challenges for rigorous NIAH RAG applications. The results advocate for cautious deployment of deliberative reasoning models in production RAG systems, necessitating additional constraint mechanisms and verification protocols to ensure reliable operation. 

\section{Conclusions}
\noindent
In this paper, we systematically investigate the evolving relationship between LLMs and RAG through the lens of the needle-in-a-haystack paradigm. We mapped RAG to the ``finding a needle in a haystack" setting and constructed the unified framework U-NIAH. The framework extends the orginal one by incorporating multi-needle , long-context needles, and nested needle scenarios. To comprehensively assess RAG performance, we design three distinct retrieval scopes (TopK, Half-Length, and Full-Length) combined with both chronological and reverse document ordering strategies. Furthermore, we develop specialized challenge cases to stress-test RAG systems under adversarial conditions.

The experimental results show that although the long-context window and capabilities of LLMs are continuously improving, RAG can still significantly enhance the performance of LLMs overall. However, how to better utilize RAG is affected by many dimensions. Retrieving more chunks does not enhance the effect of RAG, but instead introduces more noise and significantly increases the proportion of hallucinations generated by LLMs. Moreover, the ordering of retrieved documents also affects the final results. Arranging chunks in reverse order from low to high similarity, although the most critical information will be closer to the query, it may also 
increasing omission risks

This paper also tests the Reasoning LLMs that are currently becoming the main development direction. They indeed show better performance than conventional LLMs in the NIAH tasks of ICL. But they are not suitable for the RAG scenario. Excessive thinking is more likely to introduce external distraction and make LLMs more susceptible to noise interference. When there is semantic interference between the background and key information, RAG on reasoning models is even worse than direct answering. This poses a challenge to how to utilize these powerful thinking models in RAG. In future research, we will further explore optimizing the performance of reasoning models in long-context scenarios through more appropriate retrieval processes, such as iterative retrieval and active retrieval.

\bibliographystyle{ACM-Reference-Format}
\bibliography{u-niah}


\begin{thebibliography}{33}

%%% the Web address.

\ifx \showCODEN    \undefined \def \showCODEN     #1{\unskip}     \fi
\ifx \showISBNx    \undefined \def \showISBNx     #1{\unskip}     \fi
\ifx \showISBNxiii \undefined \def \showISBNxiii  #1{\unskip}     \fi
\ifx \showISSN     \undefined \def \showISSN      #1{\unskip}     \fi
\ifx \showLCCN     \undefined \def \showLCCN      #1{\unskip}     \fi
\ifx \shownote     \undefined \def \shownote      #1{#1}          \fi
\ifx \showarticletitle \undefined \def \showarticletitle #1{#1}   \fi
\ifx \showURL      \undefined \def \showURL       {\relax}        \fi
% invisible to TeX
\providecommand\bibfield[2]{#2}
\providecommand\bibinfo[2]{#2}
\providecommand\natexlab[1]{#1}
\providecommand\showeprint[2][]{arXiv:#2}

\bibitem[Bai et~al\mbox{.}(2024a)]%
        {bai-etal-2024-longbench}
\bibfield{author}{\bibinfo{person}{Yushi Bai}, \bibinfo{person}{Xin Lv}, \bibinfo{person}{Jiajie Zhang}, \bibinfo{person}{Hongchang Lyu}, \bibinfo{person}{Jiankai Tang}, \bibinfo{person}{Zhidian Huang}, \bibinfo{person}{Zhengxiao Du}, \bibinfo{person}{Xiao Liu}, \bibinfo{person}{Aohan Zeng}, \bibinfo{person}{Lei Hou}, \bibinfo{person}{Yuxiao Dong}, \bibinfo{person}{Jie Tang}, {and} \bibinfo{person}{Juanzi Li}.} \bibinfo{year}{2024}\natexlab{a}.
\newblock \showarticletitle{{L}ong{B}ench: A Bilingual, Multitask Benchmark for Long Context Understanding}. In \bibinfo{booktitle}{\emph{Proceedings of the 62nd Annual Meeting of the Association for Computational Linguistics (Volume 1: Long Papers)}}, \bibfield{editor}{\bibinfo{person}{Lun-Wei Ku}, \bibinfo{person}{Andre Martins}, {and} \bibinfo{person}{Vivek Srikumar}} (Eds.). \bibinfo{publisher}{Association for Computational Linguistics}, \bibinfo{address}{Bangkok, Thailand}, \bibinfo{pages}{3119--3137}.
\newblock
\href{https://doi.org/10.18653/v1/2024.acl-long.172}{doi:\nolinkurl{10.18653/v1/2024.acl-long.172}}


\bibitem[Bai et~al\mbox{.}(2024b)]%
        {bai2024longbenchv2}
\bibfield{author}{\bibinfo{person}{Yushi Bai}, \bibinfo{person}{Shangqing Tu}, \bibinfo{person}{Jiajie Zhang}, \bibinfo{person}{Hao Peng}, \bibinfo{person}{Xiaozhi Wang}, \bibinfo{person}{Xin Lv}, \bibinfo{person}{Shulin Cao}, \bibinfo{person}{Jiazheng Xu}, \bibinfo{person}{Lei Hou}, \bibinfo{person}{Yuxiao Dong}, {et~al\mbox{.}}} \bibinfo{year}{2024}\natexlab{b}.
\newblock \showarticletitle{LongBench v2: Towards deeper understanding and reasoning on realistic long-context multitasks}.
\newblock \bibinfo{journal}{\emph{arXiv preprint arXiv:2412.15204}} (\bibinfo{year}{2024}).
\newblock


\bibitem[Basmov et~al\mbox{.}(2024)]%
        {basmov2024llms}
\bibfield{author}{\bibinfo{person}{Victoria Basmov}, \bibinfo{person}{Yoav Goldberg}, {and} \bibinfo{person}{Reut Tsarfaty}.} \bibinfo{year}{2024}\natexlab{}.
\newblock \showarticletitle{LLMs' Reading Comprehension Is Affected by Parametric Knowledge and Struggles with Hypothetical Statements}.
\newblock \bibinfo{journal}{\emph{arXiv preprint arXiv:2404.06283}} (\bibinfo{year}{2024}).
\newblock


\bibitem[Bulatov et~al\mbox{.}(2023)]%
        {bulatov2023scaling}
\bibfield{author}{\bibinfo{person}{Aydar Bulatov}, \bibinfo{person}{Yuri Kuratov}, \bibinfo{person}{Yermek Kapushev}, {and} \bibinfo{person}{Mikhail~S Burtsev}.} \bibinfo{year}{2023}\natexlab{}.
\newblock \showarticletitle{Scaling transformer to 1m tokens and beyond with rmt}.
\newblock \bibinfo{journal}{\emph{arXiv preprint arXiv:2304.11062}} (\bibinfo{year}{2023}).
\newblock


\bibitem[Chen et~al\mbox{.}(2023)]%
        {chen2023extending}
\bibfield{author}{\bibinfo{person}{Shouyuan Chen}, \bibinfo{person}{Sherman Wong}, \bibinfo{person}{Liangjian Chen}, {and} \bibinfo{person}{Yuandong Tian}.} \bibinfo{year}{2023}\natexlab{}.
\newblock \showarticletitle{Extending context window of large language models via positional interpolation}.
\newblock \bibinfo{journal}{\emph{arXiv preprint arXiv:2306.15595}} (\bibinfo{year}{2023}).
\newblock


\bibitem[Chen et~al\mbox{.}(2024)]%
        {longlora}
\bibfield{author}{\bibinfo{person}{Yukang Chen}, \bibinfo{person}{Shengju Qian}, \bibinfo{person}{Haotian Tang}, \bibinfo{person}{Xin Lai}, \bibinfo{person}{Zhijian Liu}, \bibinfo{person}{Song Han}, {and} \bibinfo{person}{Jiaya Jia}.} \bibinfo{year}{2024}\natexlab{}.
\newblock \showarticletitle{LongLoRA: Efficient Fine-tuning of Long-Context Large Language Models}. In \bibinfo{booktitle}{\emph{The International Conference on Learning Representations (ICLR)}}.
\newblock


\bibitem[Fei et~al\mbox{.}(2024)]%
        {fei-etal-2024-extending}
\bibfield{author}{\bibinfo{person}{Weizhi Fei}, \bibinfo{person}{Xueyan Niu}, \bibinfo{person}{Pingyi Zhou}, \bibinfo{person}{Lu Hou}, \bibinfo{person}{Bo Bai}, \bibinfo{person}{Lei Deng}, {and} \bibinfo{person}{Wei Han}.} \bibinfo{year}{2024}\natexlab{}.
\newblock \showarticletitle{Extending Context Window of Large Language Models via Semantic Compression}. In \bibinfo{booktitle}{\emph{Findings of the Association for Computational Linguistics: ACL 2024}}, \bibfield{editor}{\bibinfo{person}{Lun-Wei Ku}, \bibinfo{person}{Andre Martins}, {and} \bibinfo{person}{Vivek Srikumar}} (Eds.). \bibinfo{publisher}{Association for Computational Linguistics}, \bibinfo{address}{Bangkok, Thailand}, \bibinfo{pages}{5169--5181}.
\newblock
\href{https://doi.org/10.18653/v1/2024.findings-acl.306}{doi:\nolinkurl{10.18653/v1/2024.findings-acl.306}}


\bibitem[Gao et~al\mbox{.}(2023)]%
        {retrieval_survey}
\bibfield{author}{\bibinfo{person}{Yunfan Gao}, \bibinfo{person}{Yun Xiong}, \bibinfo{person}{Xinyu Gao}, \bibinfo{person}{Kangxiang Jia}, \bibinfo{person}{Jinliu Pan}, \bibinfo{person}{Yuxi Bi}, \bibinfo{person}{Yi Dai}, \bibinfo{person}{Jiawei Sun}, {and} \bibinfo{person}{Haofen Wang}.} \bibinfo{year}{2023}\natexlab{}.
\newblock \showarticletitle{Retrieval-augmented generation for large language models: A survey}.
\newblock \bibinfo{journal}{\emph{arXiv preprint arXiv:2312.10997}} (\bibinfo{year}{2023}).
\newblock


\bibitem[Gao et~al\mbox{.}(2024)]%
        {gao2024modular}
\bibfield{author}{\bibinfo{person}{Yunfan Gao}, \bibinfo{person}{Yun Xiong}, \bibinfo{person}{Meng Wang}, {and} \bibinfo{person}{Haofen Wang}.} \bibinfo{year}{2024}\natexlab{}.
\newblock \showarticletitle{Modular rag: Transforming rag systems into lego-like reconfigurable frameworks}.
\newblock \bibinfo{journal}{\emph{arXiv preprint arXiv:2407.21059}} (\bibinfo{year}{2024}).
\newblock


\bibitem[Guo et~al\mbox{.}(2025)]%
        {deepseek_r1}
\bibfield{author}{\bibinfo{person}{Daya Guo}, \bibinfo{person}{Dejian Yang}, \bibinfo{person}{Haowei Zhang}, \bibinfo{person}{Junxiao Song}, \bibinfo{person}{Ruoyu Zhang}, \bibinfo{person}{Runxin Xu}, \bibinfo{person}{Qihao Zhu}, \bibinfo{person}{Shirong Ma}, \bibinfo{person}{Peiyi Wang}, \bibinfo{person}{Xiao Bi}, {et~al\mbox{.}}} \bibinfo{year}{2025}\natexlab{}.
\newblock \showarticletitle{Deepseek-r1: Incentivizing reasoning capability in llms via reinforcement learning}.
\newblock \bibinfo{journal}{\emph{arXiv preprint arXiv:2501.12948}} (\bibinfo{year}{2025}).
\newblock


\bibitem[Hsieh et~al\mbox{.}(2024)]%
        {hsieh2024ruler}
\bibfield{author}{\bibinfo{person}{Cheng-Ping Hsieh}, \bibinfo{person}{Simeng Sun}, \bibinfo{person}{Samuel Kriman}, \bibinfo{person}{Shantanu Acharya}, \bibinfo{person}{Dima Rekesh}, \bibinfo{person}{Fei Jia}, \bibinfo{person}{Yang Zhang}, {and} \bibinfo{person}{Boris Ginsburg}.} \bibinfo{year}{2024}\natexlab{}.
\newblock \showarticletitle{RULER: What's the Real Context Size of Your Long-Context Language Models?}
\newblock \bibinfo{journal}{\emph{arXiv preprint arXiv:2404.06654}} (\bibinfo{year}{2024}).
\newblock


\bibitem[Huang et~al\mbox{.}(2023)]%
        {long-context_survey}
\bibfield{author}{\bibinfo{person}{Yunpeng Huang}, \bibinfo{person}{Jingwei Xu}, \bibinfo{person}{Junyu Lai}, \bibinfo{person}{Zixu Jiang}, \bibinfo{person}{Taolue Chen}, \bibinfo{person}{Zenan Li}, \bibinfo{person}{Yuan Yao}, \bibinfo{person}{Xiaoxing Ma}, \bibinfo{person}{Lijuan Yang}, \bibinfo{person}{Hao Chen}, {et~al\mbox{.}}} \bibinfo{year}{2023}\natexlab{}.
\newblock \showarticletitle{Advancing transformer architecture in long-context large language models: A comprehensive survey}.
\newblock \bibinfo{journal}{\emph{arXiv preprint arXiv:2311.12351}} (\bibinfo{year}{2023}).
\newblock


\bibitem[Jin et~al\mbox{.}(2024)]%
        {jin2024long}
\bibfield{author}{\bibinfo{person}{Bowen Jin}, \bibinfo{person}{Jinsung Yoon}, \bibinfo{person}{Jiawei Han}, {and} \bibinfo{person}{Sercan~O Arik}.} \bibinfo{year}{2024}\natexlab{}.
\newblock \showarticletitle{Long-context llms meet rag: Overcoming challenges for long inputs in rag}.
\newblock \bibinfo{journal}{\emph{arXiv preprint arXiv:2410.05983}} (\bibinfo{year}{2024}).
\newblock


\bibitem[Laban et~al\mbox{.}(2024)]%
        {laban2024summary}
\bibfield{author}{\bibinfo{person}{Philippe Laban}, \bibinfo{person}{Alexander~R Fabbri}, \bibinfo{person}{Caiming Xiong}, {and} \bibinfo{person}{Chien-Sheng Wu}.} \bibinfo{year}{2024}\natexlab{}.
\newblock \showarticletitle{Summary of a haystack: A challenge to long-context llms and rag systems}.
\newblock \bibinfo{journal}{\emph{arXiv preprint arXiv:2407.01370}} (\bibinfo{year}{2024}).
\newblock


\bibitem[Leng et~al\mbox{.}(2024)]%
        {leng2024long}
\bibfield{author}{\bibinfo{person}{Quinn Leng}, \bibinfo{person}{Jacob Portes}, \bibinfo{person}{Sam Havens}, \bibinfo{person}{Matei Zaharia}, {and} \bibinfo{person}{Michael Carbin}.} \bibinfo{year}{2024}\natexlab{}.
\newblock \showarticletitle{Long Context RAG Performance of Large Language Models}. In \bibinfo{booktitle}{\emph{Advances in Neural Information Processing Systems (NeurIPS) Workshops: Track on Adaptive Foundation Models: Evolving AI for Personalized and Efficient Learning}} \emph{(\bibinfo{series}{NeurIPS'24 Workshops})}.
\newblock
\urldef\tempurl%
\url{https://arxiv.org/abs/2411.03538}
\showURL{%
\tempurl}
\newblock
\shownote{Workshop paper at NeurIPS 2024}.


\bibitem[Li et~al\mbox{.}(2024a)]%
        {li2024long}
\bibfield{author}{\bibinfo{person}{Xinze Li}, \bibinfo{person}{Yixin Cao}, \bibinfo{person}{Yubo Ma}, {and} \bibinfo{person}{Aixin Sun}.} \bibinfo{year}{2024}\natexlab{a}.
\newblock \showarticletitle{Long Context vs. RAG for LLMs: An Evaluation and Revisits}.
\newblock \bibinfo{journal}{\emph{arXiv preprint arXiv:2501.01880}} (\bibinfo{year}{2024}).
\newblock


\bibitem[Li et~al\mbox{.}(2024b)]%
        {li-etal-2024-retrieval}
\bibfield{author}{\bibinfo{person}{Zhuowan Li}, \bibinfo{person}{Cheng Li}, \bibinfo{person}{Mingyang Zhang}, \bibinfo{person}{Qiaozhu Mei}, {and} \bibinfo{person}{Michael Bendersky}.} \bibinfo{year}{2024}\natexlab{b}.
\newblock \showarticletitle{Retrieval Augmented Generation or Long-Context {LLM}s? A Comprehensive Study and Hybrid Approach}. In \bibinfo{booktitle}{\emph{Proceedings of the 2024 Conference on Empirical Methods in Natural Language Processing: Industry Track}}, \bibfield{editor}{\bibinfo{person}{Franck Dernoncourt}, \bibinfo{person}{Daniel Preo{\c{t}}iuc-Pietro}, {and} \bibinfo{person}{Anastasia Shimorina}} (Eds.). \bibinfo{publisher}{Association for Computational Linguistics}, \bibinfo{address}{Miami, Florida, US}, \bibinfo{pages}{881--893}.
\newblock
\href{https://doi.org/10.18653/v1/2024.emnlp-industry.66}{doi:\nolinkurl{10.18653/v1/2024.emnlp-industry.66}}


\bibitem[Liu et~al\mbox{.}(2024a)]%
        {deepseek_v3}
\bibfield{author}{\bibinfo{person}{Aixin Liu}, \bibinfo{person}{Bei Feng}, \bibinfo{person}{Bing Xue}, \bibinfo{person}{Bingxuan Wang}, \bibinfo{person}{Bochao Wu}, \bibinfo{person}{Chengda Lu}, \bibinfo{person}{Chenggang Zhao}, \bibinfo{person}{Chengqi Deng}, \bibinfo{person}{Chenyu Zhang}, \bibinfo{person}{Chong Ruan}, {et~al\mbox{.}}} \bibinfo{year}{2024}\natexlab{a}.
\newblock \showarticletitle{Deepseek-v3 technical report}.
\newblock \bibinfo{journal}{\emph{arXiv preprint arXiv:2412.19437}} (\bibinfo{year}{2024}).
\newblock


\bibitem[Liu et~al\mbox{.}(2024b)]%
        {lost_in_the_mddile}
\bibfield{author}{\bibinfo{person}{Nelson~F Liu}, \bibinfo{person}{Kevin Lin}, \bibinfo{person}{John Hewitt}, \bibinfo{person}{Ashwin Paranjape}, \bibinfo{person}{Michele Bevilacqua}, \bibinfo{person}{Fabio Petroni}, {and} \bibinfo{person}{Percy Liang}.} \bibinfo{year}{2024}\natexlab{b}.
\newblock \showarticletitle{Lost in the middle: How language models use long contexts}.
\newblock \bibinfo{journal}{\emph{Transactions of the Association for Computational Linguistics}}  \bibinfo{volume}{12} (\bibinfo{year}{2024}), \bibinfo{pages}{157--173}.
\newblock


\bibitem[Qiu et~al\mbox{.}(2025)]%
        {qiu2025eliciting}
\bibfield{author}{\bibinfo{person}{Yifu Qiu}, \bibinfo{person}{Varun Embar}, \bibinfo{person}{Yizhe Zhang}, \bibinfo{person}{Navdeep Jaitly}, \bibinfo{person}{Shay~B Cohen}, {and} \bibinfo{person}{Benjamin Han}.} \bibinfo{year}{2025}\natexlab{}.
\newblock \showarticletitle{Eliciting In-context Retrieval and Reasoning for Long-context Large Language Models}.
\newblock \bibinfo{journal}{\emph{arXiv preprint arXiv:2501.08248}} (\bibinfo{year}{2025}).
\newblock


\bibitem[Rein et~al\mbox{.}(2023)]%
        {gpqa}
\bibfield{author}{\bibinfo{person}{David Rein}, \bibinfo{person}{Betty~Li Hou}, \bibinfo{person}{Asa~Cooper Stickland}, \bibinfo{person}{Jackson Petty}, \bibinfo{person}{Richard~Yuanzhe Pang}, \bibinfo{person}{Julien Dirani}, \bibinfo{person}{Julian Michael}, {and} \bibinfo{person}{Samuel~R Bowman}.} \bibinfo{year}{2023}\natexlab{}.
\newblock \showarticletitle{Gpqa: A graduate-level google-proof q\&a benchmark}.
\newblock \bibinfo{journal}{\emph{arXiv preprint arXiv:2311.12022}} (\bibinfo{year}{2023}).
\newblock


\bibitem[Saad-Falcon et~al\mbox{.}(2024)]%
        {saad2024benchmarking}
\bibfield{author}{\bibinfo{person}{Jon Saad-Falcon}, \bibinfo{person}{Daniel~Y Fu}, \bibinfo{person}{Simran Arora}, \bibinfo{person}{Neel Guha}, {and} \bibinfo{person}{Christopher R{\'e}}.} \bibinfo{year}{2024}\natexlab{}.
\newblock \showarticletitle{Benchmarking and building long-context retrieval models with loco and m2-bert}.
\newblock \bibinfo{journal}{\emph{arXiv preprint arXiv:2402.07440}} (\bibinfo{year}{2024}).
\newblock


\bibitem[Wang et~al\mbox{.}(2024a)]%
        {wang-etal-2024-leave}
\bibfield{author}{\bibinfo{person}{Minzheng Wang}, \bibinfo{person}{Longze Chen}, \bibinfo{person}{Fu Cheng}, \bibinfo{person}{Shengyi Liao}, \bibinfo{person}{Xinghua Zhang}, \bibinfo{person}{Bingli Wu}, \bibinfo{person}{Haiyang Yu}, \bibinfo{person}{Nan Xu}, \bibinfo{person}{Lei Zhang}, \bibinfo{person}{Run Luo}, \bibinfo{person}{Yunshui Li}, \bibinfo{person}{Min Yang}, \bibinfo{person}{Fei Huang}, {and} \bibinfo{person}{Yongbin Li}.} \bibinfo{year}{2024}\natexlab{a}.
\newblock \showarticletitle{Leave No Document Behind: Benchmarking Long-Context {LLM}s with Extended Multi-Doc {QA}}. In \bibinfo{booktitle}{\emph{Proceedings of the 2024 Conference on Empirical Methods in Natural Language Processing}}, \bibfield{editor}{\bibinfo{person}{Yaser Al-Onaizan}, \bibinfo{person}{Mohit Bansal}, {and} \bibinfo{person}{Yun-Nung Chen}} (Eds.). \bibinfo{publisher}{Association for Computational Linguistics}, \bibinfo{address}{Miami, Florida, USA}, \bibinfo{pages}{5627--5646}.
\newblock
\href{https://doi.org/10.18653/v1/2024.emnlp-main.322}{doi:\nolinkurl{10.18653/v1/2024.emnlp-main.322}}


\bibitem[Wang et~al\mbox{.}(2024b)]%
        {ijcai2024p917}
\bibfield{author}{\bibinfo{person}{Xindi Wang}, \bibinfo{person}{Mahsa Salmani}, \bibinfo{person}{Parsa Omidi}, \bibinfo{person}{Xiangyu Ren}, \bibinfo{person}{Mehdi Rezagholizadeh}, {and} \bibinfo{person}{Armaghan Eshaghi}.} \bibinfo{year}{2024}\natexlab{b}.
\newblock \showarticletitle{Beyond the Limits: A Survey of Techniques to Extend the Context Length in Large Language Models}. In \bibinfo{booktitle}{\emph{Proceedings of the Thirty-Third International Joint Conference on Artificial Intelligence, {IJCAI-24}}}, \bibfield{editor}{\bibinfo{person}{Kate Larson}} (Ed.). \bibinfo{publisher}{International Joint Conferences on Artificial Intelligence Organization}, \bibinfo{pages}{8299--8307}.
\newblock
\href{https://doi.org/10.24963/ijcai.2024/917}{doi:\nolinkurl{10.24963/ijcai.2024/917}}
\newblock
\shownote{Survey Track}.


\bibitem[White et~al\mbox{.}(2023)]%
        {white2023prompt}
\bibfield{author}{\bibinfo{person}{Jules White}, \bibinfo{person}{Quchen Fu}, \bibinfo{person}{Sam Hays}, \bibinfo{person}{Michael Sandborn}, \bibinfo{person}{Carlos Olea}, \bibinfo{person}{Henry Gilbert}, \bibinfo{person}{Ashraf Elnashar}, \bibinfo{person}{Jesse Spencer-Smith}, {and} \bibinfo{person}{Douglas~C Schmidt}.} \bibinfo{year}{2023}\natexlab{}.
\newblock \showarticletitle{A prompt pattern catalog to enhance prompt engineering with chatgpt}.
\newblock \bibinfo{journal}{\emph{arXiv preprint arXiv:2302.11382}} (\bibinfo{year}{2023}).
\newblock


\bibitem[Xiang et~al\mbox{.}(2024)]%
        {xiang2024certifiably}
\bibfield{author}{\bibinfo{person}{Chong Xiang}, \bibinfo{person}{Tong Wu}, \bibinfo{person}{Zexuan Zhong}, \bibinfo{person}{David Wagner}, \bibinfo{person}{Danqi Chen}, {and} \bibinfo{person}{Prateek Mittal}.} \bibinfo{year}{2024}\natexlab{}.
\newblock \showarticletitle{Certifiably Robust RAG against Retrieval Corruption}.
\newblock \bibinfo{journal}{\emph{arXiv preprint arXiv:2405.15556}} (\bibinfo{year}{2024}).
\newblock


\bibitem[Xu et~al\mbox{.}(2023)]%
        {xu2023retrieval}
\bibfield{author}{\bibinfo{person}{Peng Xu}, \bibinfo{person}{Wei Ping}, \bibinfo{person}{Xianchao Wu}, \bibinfo{person}{Lawrence McAfee}, \bibinfo{person}{Chen Zhu}, \bibinfo{person}{Zihan Liu}, \bibinfo{person}{Sandeep Subramanian}, \bibinfo{person}{Evelina Bakhturina}, \bibinfo{person}{Mohammad Shoeybi}, {and} \bibinfo{person}{Bryan Catanzaro}.} \bibinfo{year}{2023}\natexlab{}.
\newblock \showarticletitle{Retrieval meets long context large language models}.
\newblock \bibinfo{journal}{\emph{arXiv preprint arXiv:2310.03025}} (\bibinfo{year}{2023}).
\newblock


\bibitem[Xu et~al\mbox{.}(2024)]%
        {xu2024chatqa}
\bibfield{author}{\bibinfo{person}{Peng Xu}, \bibinfo{person}{Wei Ping}, \bibinfo{person}{Xianchao Wu}, \bibinfo{person}{Chejian Xu}, \bibinfo{person}{Zihan Liu}, \bibinfo{person}{Mohammad Shoeybi}, {and} \bibinfo{person}{Bryan Catanzaro}.} \bibinfo{year}{2024}\natexlab{}.
\newblock \showarticletitle{Chatqa 2: Bridging the gap to proprietary llms in long context and rag capabilities}.
\newblock \bibinfo{journal}{\emph{arXiv preprint arXiv:2407.14482}} (\bibinfo{year}{2024}).
\newblock


\bibitem[Yang et~al\mbox{.}(2024)]%
        {qwen2}
\bibfield{author}{\bibinfo{person}{An Yang}, \bibinfo{person}{Baosong Yang}, \bibinfo{person}{Beichen Zhang}, \bibinfo{person}{Binyuan Hui}, \bibinfo{person}{Bo Zheng}, \bibinfo{person}{Bowen Yu}, \bibinfo{person}{Chengyuan Li}, \bibinfo{person}{Dayiheng Liu}, \bibinfo{person}{Fei Huang}, \bibinfo{person}{Haoran Wei}, {et~al\mbox{.}}} \bibinfo{year}{2024}\natexlab{}.
\newblock \showarticletitle{Qwen2. 5 technical report}.
\newblock \bibinfo{journal}{\emph{arXiv preprint arXiv:2412.15115}} (\bibinfo{year}{2024}).
\newblock


\bibitem[Yu et~al\mbox{.}(2024)]%
        {oprag}
\bibfield{author}{\bibinfo{person}{Tan Yu}, \bibinfo{person}{Anbang Xu}, {and} \bibinfo{person}{Rama Akkiraju}.} \bibinfo{year}{2024}\natexlab{}.
\newblock \showarticletitle{In defense of rag in the era of long-context language models}.
\newblock \bibinfo{journal}{\emph{arXiv preprint arXiv:2409.01666}} (\bibinfo{year}{2024}).
\newblock


\bibitem[Zhang et~al\mbox{.}(2024b)]%
        {zhang2024bench}
\bibfield{author}{\bibinfo{person}{Xinrong Zhang}, \bibinfo{person}{Yingfa Chen}, \bibinfo{person}{Shengding Hu}, \bibinfo{person}{Zihang Xu}, \bibinfo{person}{Junhao Chen}, \bibinfo{person}{Moo Hao}, \bibinfo{person}{Xu Han}, \bibinfo{person}{Zhen Thai}, \bibinfo{person}{Shuo Wang}, \bibinfo{person}{Zhiyuan Liu}, {et~al\mbox{.}}} \bibinfo{year}{2024}\natexlab{b}.
\newblock \showarticletitle{$\infty$ Bench: Extending long context evaluation beyond 100k tokens}. In \bibinfo{booktitle}{\emph{Proceedings of the 62nd Annual Meeting of the Association for Computational Linguistics (Volume 1: Long Papers)}}. \bibinfo{pages}{15262--15277}.
\newblock


\bibitem[Zhang et~al\mbox{.}(2024a)]%
        {zhang-etal-2024-bench}
\bibfield{author}{\bibinfo{person}{Xinrong Zhang}, \bibinfo{person}{Yingfa Chen}, \bibinfo{person}{Shengding Hu}, \bibinfo{person}{Zihang Xu}, \bibinfo{person}{Junhao Chen}, \bibinfo{person}{Moo Hao}, \bibinfo{person}{Xu Han}, \bibinfo{person}{Zhen Thai}, \bibinfo{person}{Shuo Wang}, \bibinfo{person}{Zhiyuan Liu}, {and} \bibinfo{person}{Maosong Sun}.} \bibinfo{year}{2024}\natexlab{a}.
\newblock \showarticletitle{$\infty${B}ench: Extending Long Context Evaluation Beyond 100{K} Tokens}. In \bibinfo{booktitle}{\emph{Proceedings of the 62nd Annual Meeting of the Association for Computational Linguistics (Volume 1: Long Papers)}}, \bibfield{editor}{\bibinfo{person}{Lun-Wei Ku}, \bibinfo{person}{Andre Martins}, {and} \bibinfo{person}{Vivek Srikumar}} (Eds.). \bibinfo{publisher}{Association for Computational Linguistics}, \bibinfo{address}{Bangkok, Thailand}, \bibinfo{pages}{15262--15277}.
\newblock
\href{https://doi.org/10.18653/v1/2024.acl-long.814}{doi:\nolinkurl{10.18653/v1/2024.acl-long.814}}


\bibitem[Zhao et~al\mbox{.}(2024)]%
        {zhao-etal-2024-longrag}
\bibfield{author}{\bibinfo{person}{Qingfei Zhao}, \bibinfo{person}{Ruobing Wang}, \bibinfo{person}{Yukuo Cen}, \bibinfo{person}{Daren Zha}, \bibinfo{person}{Shicheng Tan}, \bibinfo{person}{Yuxiao Dong}, {and} \bibinfo{person}{Jie Tang}.} \bibinfo{year}{2024}\natexlab{}.
\newblock \showarticletitle{{L}ong{RAG}: A Dual-Perspective Retrieval-Augmented Generation Paradigm for Long-Context Question Answering}. In \bibinfo{booktitle}{\emph{Proceedings of the 2024 Conference on Empirical Methods in Natural Language Processing}}, \bibfield{editor}{\bibinfo{person}{Yaser Al-Onaizan}, \bibinfo{person}{Mohit Bansal}, {and} \bibinfo{person}{Yun-Nung Chen}} (Eds.). \bibinfo{publisher}{Association for Computational Linguistics}, \bibinfo{address}{Miami, Florida, USA}, \bibinfo{pages}{22600--22632}.
\newblock
\href{https://doi.org/10.18653/v1/2024.emnlp-main.1259}{doi:\nolinkurl{10.18653/v1/2024.emnlp-main.1259}}


\end{thebibliography}

\end{document}